\documentclass[aip,cha, reprint]{revtex4-1}

\usepackage
{
    amsmath,
    amssymb,
    amsthm,
    enumitem,
    graphicx,
    mathtools,
    xcolor,
    nicefrac,
    bm,
    gensymb
}
\usepackage[utf8]{inputenc}
\usepackage[T1]{fontenc}
\usepackage{etoolbox}
\usepackage[bf]{caption}
\usepackage{subcaption}
\usepackage[normalem]{ulem}
\usepackage[ruled,vlined, linesnumbered]{algorithm2e}

\usepackage[pdffitwindow=false,
            plainpages=false,
            pdfpagelabels=true,
            pdfpagemode=UseOutlines,
            pdfpagelayout=SinglePage,
            bookmarks=false,
            colorlinks=true,
            hyperfootnotes=false,
            hypertexnames=false,
            linkcolor=blue,
            urlcolor=blue!30!black,
            citecolor=green!50!black]{hyperref}

\DeclareMathOperator{\tr}{trace}
\DeclareMathOperator{\mspan}{span}
\DeclareMathOperator*{\argmin}{arg\,min}
\newcommand{\R}{\mathbb{R}}                                      % real numbers
\newcommand{\C}{\mathbb{C}}                                      % complex numbers
\newcommand{\N}{\mathbb{N}}                                      % natural numbers
\newcommand{\K}{\mathcal{K}}                                     % Koopman operator
\renewcommand{\H}{\mathbb{H}}                                    % RKHS
\newcommand{\innerprod}[2]{\left\langle #1,\, #2 \right\rangle}  % scalar product
\newcommand{\ts}{\hspace*{0.1em}}                                % thin space
\providecommand{\norm}[1]{\left\lVert #1 \right\rVert}           % norm
\renewcommand{\vec}[1]{\bm{#1}}

\newtheorem{theorem}{Theorem}[section]

\newtheorem{definition}[theorem]{Definition}
\theoremstyle{definition}

\newtheorem{remark}[theorem]{Remark}

\begin{document}

\title{Bilevel optimization for data-driven learning of Koopman embeddings using kernel-based autoencoders}

\author{Joel-Pascal Ntwali N'konzi}
\email[]{j.n.n'konzi@sms.ed.ac.uk}
\affiliation{Maxwell Institute for Mathematical Sciences, The University of Edinburgh \& Heriot--Watt University, Edinburgh, United Kingdom}

\author{Feliks N\"{u}ske}
\email[]{nueske@mpi-magdeburg.mpg.de}
\affiliation{Max Planck Institute for Dynamics of Complex Technical Systems, Magdeburg, Germany}
\affiliation{LAAS-CNRS, Université de Toulouse, Toulouse, France}

\author{Stefan Klus}
\email[]{s.klus@hw.ac.uk}
\affiliation{School of Mathematical \& Computer Sciences, Heriot--Watt University, Edinburgh, United Kingdom}

\date{\today}

\begin{abstract}
Koopman operator theory provides a linear framework for analyzing nonlinear dynamical systems and has become a major tool for data-driven modeling. A central challenge, however, is that finite-dimensional approximations computed by methods such as extended dynamic mode decomposition (EDMD) require the dictionary to be specified \textit{a priori}. Recent machine-learning approaches address this limitation by learning the dictionary from data, predominantly using artificial neural network (ANN) autoencoder architectures. Although kernel methods offer an alternative with greater interpretability and tractability for theoretical analysis, they have received little attention in this setting. We introduce \emph{extended dynamic mode decomposition with kernel-based dictionary learning} (EDMD-kDL), a kernel-based method for learning finite-dimensional Koopman embeddings directly from data. The method combines ideas from collocation methods and bilevel optimization to simultaneously learn a kernel dictionary and the corresponding Koopman approximation. We evaluate EDMD-kDL against state-of-the-art ANN-based approaches on a range of numerical experiments, including global sea-surface-temperature forecasting and learning directly from video data. Across all tested settings, EDMD-kDL achieves performance comparable to or better than the ANN-based methods. Moreover, in contrast to standard kernel methods, the proposed approach is scalable to large datasets by design since the size of the required kernel matrices depends on the number of collocation points rather than the size of the training dataset.
\end{abstract}

\maketitle 

\begin{quotation}
The Koopman operator approach to dynamics, which provides a way of treating nonlinear dynamical systems with linear techniques, is at the forefront of the exploding field of data-driven modeling and analysis of complex systems. The recent rise in prominence is mostly due to the development of algorithms, such as extended dynamic mode decomposition (EDMD) and its variants, for obtaining finite-dimensional approximations of the Koopman operator from trajectory data. Despite its great success, the EDMD algorithm is limited by the constraint that a suitable subspace in which the operator approximation is carried out has to be determined \textit{a priori}, which is a challenging task in practice. This has prompted the development of various alternative approaches that aim to identify such subspaces directly from data using machine learning. The vast majority of these methods rely on artificial neural network (ANN)-based autoencoders, where the encoder parametrizes a mapping from the state space to a higher dimensional embedding space, which in turn determines the subspace of interest. Such mappings are then called \emph{Koopman embeddings}. Despite the fact that kernel methods can be as expressive as ANNs, and are both more interpretable and amenable to theoretical analysis, there has been no attempt to apply them to this problem. In this paper, we develop a novel kernel-based algorithm for learning finite-dimensional Koopman embeddings from data that we call \emph{extended dynamic mode decomposition with kernel-based dictionary learning} (EDMD-kDL). We derive the proposed method by leveraging ideas from collocation methods and bilevel optimization, and compare its performance against state-of-the-art ANN-based approaches using both simulated and real-world datasets.
\end{quotation}

\section{Introduction}

Dynamical systems, whether deterministic or stochastic, are ubiquitous in science and engineering due to their effectiveness in modeling a wide range of natural phenomena. In contrast to the standard geometric approach to dynamics, the Koopman perspective focuses on the temporal evolution of observables, which are scalar functions of the state. The \emph{Koopman operator}~\cite{Ko31, koopman1932dynamical, LaMa94, Mezic05} is linear, albeit infinite-dimensional, even when the state-space dynamics are nonlinear.

Despite the framework being almost a century old, its recent gain in prominence is mostly due to the development, over the past two decades, of data-driven learning algorithms that yield finite-dimensional approximations of the Koopman operator or its associated spectral information from simulation or observational data; see, e.g., Refs.~\onlinecite{Schmid10, WKR15, KKS16, LDBK17, KNKWKSN18, NoNu13}. This includes in particular the \emph{extended dynamic mode decomposition} (EDMD) algorithm~\cite{WKR15, KKS16} and the \emph{variational approach to conformation dynamics} (VAC)~\cite{NoNu13, NKPMN14}. As the name suggests, EDMD is an extension of \emph{dynamic mode decomposition} (DMD)~\cite{Schmid10}, an approach that was initially introduced for decomposing fluid flows into coherent patterns, whereas VAC generalizes \emph{time-lagged independent component analysis} (TICA)~\cite{MS94}, see Refs.~\onlinecite{mezic2013analysis, TRLBK14, KKS16, KNKWKSN18, otto2021koopman, schmid2022dynamic, colbrook2024multiverse} and references therein for more details. These advances have placed the operator approach to dynamics at the forefront of the exploding field of data-driven modeling and analysis of dynamical systems. Applications include system identification~\cite{MauGon16, bruder2019nonlinear, KNPNCS20}, model reduction~\cite{PeiKlul19, KNKWKSN18, nateghi2025kinetically}, model predictive control~\cite{KorMez16, surana2016koopman, goswami2017global, arbabi2018data, BSH21}, and time-series forecasting~\cite{HNDLG17, Gia19, AELM20, LBK21, bevanda2023koopman}.

The EDMD algorithm works by considering a dictionary of functions defining a so-called \emph{Koopman embedding}, i.e., a mapping from the state space to a typically high-dimensional space such that the embedded dynamical system is, at least approximately, linear. Then, given training data, the matrix representing the operator is estimated by solving a linear regression problem, and approximate eigenvalues and eigenfunctions of the Koopman operator are obtained from the spectral decomposition of this matrix~\cite{WKR15}.

The main limitation of standard EDMD comes from the fact that designing informative Koopman embeddings is in general challenging. The same is true for the standard VAC algorithm. This has prompted the development of various dictionary learning approaches for EDMD/VAC in the literature~\cite{LDBK17, takeishi2017learning, MPWN18, wehmeyer2018time, yeung2019learning, Otto19, AELM20, TLK25}. At the core, these methods typically consider autoencoder artificial neural networks (ANNs), where the encoder parametrizes a Koopman embedding map and the (possibly fixed) decoder provides a way of reconstructing inputs from the corresponding embedding representations. The network is then trained with a linear embedded dynamics objective in addition to the standard encoder--decoder loss. Differences among these approaches include the use of hybrid dictionaries~\cite{LDBK17}, stability promotion~\cite{erichson2019physics}, training with a multi-step prediction loss~\cite{Otto19}, learning consistent forward and backward linear models in the embedding space~\cite{AELM20}, and the focus on stochastic systems~\cite{MPWN18, wehmeyer2018time}.

An alternative approach for tackling the dictionary selection problem is \emph{kernel EDMD}~\cite{WRK15, KSM19}, which has been derived in two separate ways: the first strategy is to apply the well-known \emph{kernel trick} to a dual formulation of the standard EDMD algorithm~\cite{WRK15}, whereas the second approach considers an empirical estimation of the Koopman operator defined on a \emph{reproducing kernel Hilbert space} (RKHS)~\cite{KSM19}. The distinguishing feature of this approach is that the embedding map can, in theory, be infinite-dimensional since only inner products between embedded vectors -- which can be obtained by evaluating a symmetric positive semi-definite kernel -- are required. In practice, however, one is constrained to work in the finite-dimensional subspace spanned by the kernel centered at the training data points.

The main theoretical underpinning of ANN-based dictionary learning for EDMD is the \emph{universal approximation theorem}~\cite{cybenko1989approximation}, which guarantees that any multivariate continuous function defined on a hypercube can be uniformly approximated with arbitrary accuracy by an appropriate ANN. However, this property is not specific to ANNs. In fact, the RKHSs associated with so-called \emph{universal kernels} are dense in the space of continuous functions defined on compact domains with respect to the uniform norm~\cite{micchelli2006universal, sriperumbudur2011universality}. This prompts one to wonder whether kernel methods can be as effective as ANN-based approaches for dictionary learning within EDMD/VAC. In this work, we show that this is indeed the case by developing a novel kernel-based approach for learning finite-dimensional Koopman embeddings from data, with a focus on EDMD for deterministic dynamical systems. To achieve this, we introduce the concept of \emph{kernel-based autoencoders} by leveraging ideas from collocation methods~\cite{schaback2006kernel, chen2021solving} and bilevel optimization. The resulting method, which we call \emph{extended dynamic mode decomposition with kernel-based dictionary learning} (EDMD-kDL), is less parametrized than ANN-based counterparts and relies on numerically solving a nonlinear optimization problem involving the values taken by the Koopman embedding map at a finite set of collocation points. Furthermore, we compare EDMD-kDL against state-of-the-art ANN-based approaches on a wide range of numerical experiments using both synthetic and real-world data. We find that, across settings, EDMD-kDL either outperforms or has comparable performance to ANN-based methods.

The rest of this paper is organized as follows. In Section~\ref{sec:background} we provide the necessary background material on Koopman learning. This is followed by detailed derivations of the proposed method in Section~\ref{sec:edmd-kdl}. We consider both the case of hybrid dictionaries, where one part of the dictionary is fixed whilst the other is learned from data, and the case of fully learned dictionaries. Moreover, we show how the approach can be easily modified to use a multi-step prediction loss for training, which is not possible for standard kernel EDMD. We then provide results from extensive numerical experiments in Section~\ref{sec:numerical-experiments} and conclude with a discussion of open problems and ideas for future work in Section~\ref{sec:discussion}.

%%%%%%%%%%%%%%%%%%%%%%%%%  BACKGROUND ON KOOPMAN LEARNING  %%%%%%%%%%%%%%%%%%%%%%%%%
\section{Background on Koopman learning}\label{sec:background}

\subsection{Koopman operator theory}

In this work, we consider discrete-time dynamical systems of the form
    \begin{equation}\label{eq:discrete-dynamics}
        \vec{x}_{t+1} = \vec{f}(\vec{x}_t), \quad t \in \N,
    \end{equation}
where $\vec{x}_t \in \Omega \subseteq \R^n$ is the state and $\vec{f}\colon \Omega \to \Omega$ is a nonlinear map. The \emph{Koopman operator}~\cite{Ko31} $\K$, defined by
    \begin{equation}
        \left[\K \psi\right](\vec{x}) = \psi \left(\vec{f}(\vec{x})\right), \quad \forall \vec{x} \in \Omega,
    \end{equation}
determines the time-evolution of \emph{observables} $\psi \in \mathbb{M}$, which are real- or complex-valued functions of the state. A typical choice is $\mathbb{M} = L^2(\Omega,\mu)$ for some positive measure $\mu$, see, e.g., Refs.~\onlinecite{Mezic05, WKR15}. For vector-valued observables $\vec{\psi} = \left[\psi_1, \dots, \psi_N\right]^\top \colon \Omega \to \C^N $, the Koopman operator acts element-wise, i.e., $ \K \vec{\psi}(x) = \left[\K \psi_1(x), \dots, \K \psi_N(x)\right]^\top$.

The main property of the Koopman operator is its linearity. This is the case even when $\vec{f}$ is nonlinear. As a consequence, it is amenable to spectral decomposition. In particular, the spectral information of $\K$ encodes important properties of the dynamical system such as the dominant time scales, coherent patterns, and metastability~\cite{Mezic05, RMBSH09, Schmid10, BMM12, KNKWKSN18}. \emph{Eigenfunctions} of $\K$ are observables $\varphi$ for which the equality
    \begin{equation*}
        \K \varphi = \lambda \varphi
    \end{equation*}
holds for an associated \emph{eigenvalue} $\lambda \in \C$. It follows that subspaces of $\mathbb{M}$ spanned by sets of eigenfunctions are invariant under the action of the Koopman operator. This characterization makes it useful to linearly decompose observables in terms of eigenfunctions. In particular, for the so-called \emph{full-state observable} $\vec{g}(\vec{x}) = \vec{x}$ and a set $\{(\lambda_j, \varphi_j)\}_{j\in J}$ of eigenpairs of $\K$, the \emph{Koopman mode decomposition}~\cite{Mezic05, RMBSH09} is defined as
    \begin{equation*}
       \vec{g}(\vec{x}) = \sum_{j \in J} \vec{w}_j \varphi_j(\vec{x}),
    \end{equation*}
where the vectors $\vec{w}_j \in \C^n$ are called \emph{Koopman modes}. Here, the index set $J$ is possibly infinite-dimensional. That is, depending on the system, it might not be possible to write the full-state observable as a linear combination of finitely many eigenfunctions. The relevance of this decomposition stems from the fact that, at least in theory, one can write
    \begin{align*}
        \vec{x}_{t+1} &= \vec{g}(\vec{x}_{t+1}) = [\K \vec{g}] (\vec{x}_{t}) \\
           &= \sum_{j \in J} \vec{w}_j \K \varphi_j(\vec{x}_{t}) = \sum_{j \in J} \lambda_j \vec{w}_j \varphi_j(\vec{x}_{t}),
    \end{align*}
or, more generally, for $\ell \geq 1$ time steps:
    \begin{equation*}
        \vec{x}_{t+\ell} = \sum_{j \in J} \lambda_j^{\ell} \vec{w}_j \varphi_j( \vec{x}_t).
    \end{equation*}
Despite its attractive linearity, the Koopman operator poses multiple challenges due to its infinite dimensionality such as potentially possessing a continuous spectrum~\cite{Mezic05,colbrook2024multiverse}. From a data-driven perspective, the main interest is to obtain a finite dimensional approximation of the Koopman operator from trajectory data. Such an approximation can then be used for prediction and to obtain information on the dominant modes and associated timescales of the system. We describe various methods that have been developed for this purpose in what follows.

\subsection{Extended dynamic mode decomposition (EDMD)}
The EDMD algorithm~\cite{WKR15} is a data-driven approach that provides a finite-dimensional approximation of $\K$, its eigenvalues, eigenfunctions, and the associated Koopman modes based on data of the form $\left\{\left(\vec{x}^{(i)}, \vec{y}^{(i)} \right)\right\}_{i=1}^m$, where each pair is such that $\vec{y}^{(i)} = \vec{f}\big(\vec{x}^{(i)}\big)$. The dataset is typically arranged into two data matrices
    \begin{equation}\label{eq:edmd-data-matrices}
        X = \begin{bmatrix}
            \vec{x}^{(1)} & \dots & \vec{x}^{(m)}
        \end{bmatrix}, \quad Y = \begin{bmatrix}
            \vec{y}^{(1)} & \dots & \vec{y}^{(m)}
        \end{bmatrix},
    \end{equation}
where $Y$ is the one-step forward version of $X$. This includes data coming from a single or multiple trajectories.

Let $\mathbb{D} = \{\psi_j\}_{j=1}^N$ be a dictionary of real-valued observables and define the associated \emph{embedding map} as $\vec{x} \mapsto \vec{\psi}(\vec{x}) = \left[\psi_1(\vec{x}), \dots, \psi_N(\vec{x}) \right]^\top \in \R^N$ with $N \geq n$. In practice, it is required that the full-state observable is contained in $\mathbb{V} = \mspan \mathbb{D}  \subset \mathbb{M} $, the linear span of the dictionary. If $\mathbb{V}$ is Koopman-invariant, the dynamics in the embedded space is linear, i.e.,
    \begin{equation*}
        \vec{\psi}(\vec{x}_{t+1}) = K \vec{\psi}(\vec{x}_{t}), \quad t \in \N,
    \end{equation*}
for some matrix $K \in \R^{N \times N}$. Hence, the matrix $K$ fully determines the action of the Koopman operator $\K$ restricted to the subspace $\mathbb{V}$. EDMD then finds an estimate $\widehat{K}$ of $K$ from data by solving the linear regression problem
    \begin{equation}\label{eq:edmd-optimization}
        \widehat{K} = \underset{K \in \R^{N\times N}}{\argmin}\ts \frac{1}{m} \sum_{i=1}^m \norm{\vec{\psi}\big(\vec{y}^{(i)}\big) - K \vec{\psi}\big(\vec{x}^{(i)}\big)}_2^2.
    \end{equation}
Defining the transformed data matrices
    \begin{align*}
        \Psi_X &= \begin{bmatrix} \vec{\psi}\big(\vec{x}^{(1)}\big) & \dots & \vec{\psi}\big(\vec{x}^{(m)}\big) \end{bmatrix}, \\
         \Psi_Y &= \begin{bmatrix} \vec{\psi}\big(\vec{y}^{(1)}\big) & \dots & \vec{\psi}\big(\vec{y}^{(m)}\big)\end{bmatrix},
    \end{align*}
the (minimum-norm) solution of \eqref{eq:edmd-optimization} is given by
    \begin{equation}\label{eq:edmd-solution}
        \widehat{K} = \Psi_Y \Psi_X^\dagger = \Psi_Y \Psi_X^\top \left(\Psi_X \Psi_X^\top\right)^\dagger,
    \end{equation}
where $^\dagger$ denotes the Moore--Penrose pseudoinverse. Moreover, EDMD provides up to $N$ approximate eigenpairs $\{(\lambda_j, \varphi_j)\}_{j=1}^N$ of $\K$ with
    \begin{equation*}
        \varphi_j(x) = \vec{v}_j^\top \vec{\psi}(x),
    \end{equation*}
where $\vec{v}_j$ is a left eigenvector of $\widehat{K}$ associated with the eigenvalue $\lambda_j$. Koopman modes can similarly be approximated, see Ref.~\onlinecite{WKR15} for more details. It has been shown that the EDMD matrix $\widehat{K}$ converges to the Galerkin projection of $\K$ onto $\mathbb{V}$ in the infinite-data limit~\cite{WKR15, KKS16}.

\subsection{Reproducing kernel Hilbert spaces and kernel EDMD}

As already mentioned above, the choice of dictionary has a significant impact on the performance of the EDMD algorithm. This is due to the assumption that the dictionary spans, at least approximately, a Koopman-invariant subspace. However, handcrafting such dictionaries in practice has proven to be a challenging task. The problem becomes even more severe in high‑dimensional settings, where even seemingly simple dictionaries -- such as those constituted of all polynomials up to a fixed degree -- explode combinatorially with the dimension of the state space. To alleviate this issue, the authors of Ref.~\onlinecite{WRK15} proposed to leverage the well-known \emph{kernel trick} to perform EDMD with a possibly infinite-dimensional dictionary that is implicitly defined by a chosen kernel. The resulting method is called \emph{kernel} EDMD. We summarize this approach below using the more general derivation of Ref.~\onlinecite{KSM19}, but we need to introduce the required concepts first.

\begin{definition}[Positive semi-definite kernel]
 Let $\mathbb{X}$ be a nonempty set. A function $k\colon \mathbb{X} \times \mathbb{X} \to \R$ is a positive semi-definite (PSD) kernel if
    \begin{enumerate}
        \item $k$ is symmetric, i.e., it holds that $k(x,y) = k(y,x)$ for all $x, y \in \mathbb{X}$;
        \item for any finite number of points $x^{(1)}, \dots, x^{(m)} \in \mathbb{X}$, the Gram matrix
            \begin{equation*}
                G = \begin{bmatrix}
                    k\big(x^{(1)}, x^{(1)}\big) & \dots & k\big(x^{(1)}, x^{(m)}\big) \\
                    \vdots & \ddots & \vdots \\
                    k\big(x^{(m)}, x^{(1)}\big) & \dots & k\big(x^{(m)}, x^{(m)}\big)
                \end{bmatrix}
            \end{equation*}
        is symmetric PSD, i.e., $\vec{c}^\top G \vec{c} \geq 0$ for all $\vec{c} \in \R^m$ and for all $m \in \N$.
    \end{enumerate}
\end{definition}

\begin{definition}[Reproducing kernel Hilbert space~\cite{Schoe01, Steinwart2008:SVM}] Let $\mathbb{X}$ be a nonempty set and $\H$ a Hilbert space of functions $f\colon \mathbb{X} \to \R$. Then $\H$ is called a \emph{reproducing kernel Hilbert space} (RKHS) with inner product $\innerprod{\ts\cdot\ts}{\ts\cdot\ts}$ and induced norm $\norm{f} \coloneqq \sqrt{\innerprod{f}{f}}$ if there exists a function $k\colon \mathbb{X} \times \mathbb{X} \to \R$, called reproducing kernel, with the following properties:
    \begin{enumerate}
        \item $k$ has the reproducing property
            \begin{equation*}
                \innerprod{f}{k(x, \ts\cdot\ts)} = f(x) \text{ for all } f \in \H;
            \end{equation*}
            in particular,
            \begin{equation*}
                \innerprod{k(x, \ts\cdot\ts)}{k(x', \ts\cdot\ts)} = k(x, x').
            \end{equation*}
        \item $k$ spans $\H$, i.e., $\H = \overline{\mspan \{k(x, \ts\cdot\ts) \mid x \in \H\}}$, where $\overline{\mathbb{A}}$ is denotes the completion of the metric space $\mathbb{A}$.
    \end{enumerate}
\end{definition}
It follows from the Moore--Aronszajn theorem that every PSD kernel $k$ uniquely defines an RKHS for which it is the reproducing kernel; and, conversely, every reproducing kernel is positive semi-definite~\cite{Aronszajn50}. Henceforth, we assume that all kernels are PSD. The mapping $x \mapsto k(x, \ts\cdot\ts)$ is typically referred to as the \emph{canonical feature map}.

Consider a kernel $k$ and let $\H$ be the associated RKHS. Assuming that $\H$ is Koopman-invariant, the authors of Ref.~\onlinecite{KSM19} showed that the restriction of $\K$ onto $\H$, which they called \emph{kernel Koopman operator}, is given by
    \begin{equation}
        \K_k = \left(\mathcal{C}_{00} + \epsilon \ts \mathcal{I}\right)^{-1} \mathcal{C}_{01},
    \end{equation}
where $\epsilon$ is a positive regularization parameter, $\mathcal{I}$ is the identity operator, $\mathcal{C}_{00}$ is the covariance operator between two observables at time $t$, and $\mathcal{C}_{01}$ is the cross-covariance operator between an observable evaluated at time $t$ and another observable at time $t+1$; see Ref.~\onlinecite{KSM19} for the precise mathematical definitions. Then, for the data matrices $X$ and $Y$ as in \eqref{eq:edmd-data-matrices}, it was shown that the empirical estimate is provided by the finite-rank operator
    \begin{equation}
        \widehat{\K}_k = \vec{k}_X^\top \left(G_{XX} + m \epsilon I \right)^{-1} \vec{k}_Y,
    \end{equation}
and approximate eigenfunctions can be obtained by solving the eigenvalue problem
    \begin{equation}
        \left(G_{XX} + m \epsilon I \right)^{-1} G_{YX} \vec{v} = \lambda \vec{v},
    \end{equation}
where $G_{XX} = \left(k\big(\vec{x}^{(i)}, \vec{x}^{(j)}\big)\right)_{i,j=1}^m$ and $G_{YX} = \left(k\big(\vec{y}^{(i)}, \vec{x}^{(j)}\big)\right)_{i,j=1}^m$ are kernel matrices, and
    \begin{equation*}
        \vec{k}_X = \begin{bmatrix}
            k(\vec{x}^{(1)}, \ts\cdot\ts) \\
            \vdots \\ k(\vec{x}^{(m)}, \ts\cdot\ts)
        \end{bmatrix}, \quad
        \vec{k}_Y = \begin{bmatrix}
            k(\vec{y}^{(1)}, \ts\cdot\ts) \\ \vdots\\ k(\vec{y}^{(m)}, \ts\cdot\ts)
        \end{bmatrix}.
    \end{equation*}

\subsection{Learning Koopman dictionaries with neural networks}

Kernel EDMD relies on the assumption that the induced RKHS is Koopman-invariant. However, this might not hold in practice. An alternative approach is to simultaneously learn an approximately Koopman-invariant dictionary and the corresponding EDMD matrix directly from data. This idea was first proposed by the authors of Ref.~\onlinecite{LDBK17}, who considered hybrid dictionaries of the form
    \begin{equation}\label{eq:dictionary-with-direct-link}
        \vec{x} \mapsto \vec{\psi}_{\vec{\theta}}(\vec{x}) = \left[
            \vec{\rho}(\vec{x})^\top, \: \vec{\phi}_{\vec{\theta}}(\vec{x})^\top
        \right]^\top \in \R^N,
    \end{equation}
where $\vec{\rho} \colon \Omega \to \R^o$ is a fixed component, typically the full-state observable, and $\vec{\phi}_{\vec{\theta}} \colon \Omega \to \R^{p}$, with $N = o + p$, is a learnable complementary part parameterized by an ANN with weights and biases denoted by $\vec{\theta}$. The EDMD matrix $K$ and the ANN parameters are then optimized simultaneously by minimizing the objective function in \eqref{eq:edmd-optimization}, possibly with regularization over both $K$ and $\vec{\theta}$. Specifically, they used an iterative approach where, at each optimization step, first an exact update of $K$ for fixed $\vec{\theta}$ is made as in \eqref{eq:edmd-solution}; then $\vec{\theta}$ is updated using stochastic gradient descent for fixed $K$.

The main motivation for including a fixed component $\vec{\rho}$ was so that uninformative solutions, such as $\vec{\psi}_{\vec{\theta}}$ being constant and $K$ the identity map, could be avoided. Another approach for avoiding such trivial solutions is to consider an autoencoder ANN $\left(\vec{\psi}_{\vec{\theta}}, \vec{\varphi}_{\vec{\omega}}\right)$, where the encoder $\vec{\psi}_{\vec{\theta}}$ parametrizes the EDMD dictionary and the decoder $\vec{\varphi}_{\vec{\omega}}$ maps from the embedding to the state space~\cite{Otto19, yeung2019learning}. Then the network parameters $(\vec{\theta}, \vec{\omega})$ and the Koopman matrix $K$ are jointly optimized by minimizing a linear combination of the forward prediction loss in \eqref{eq:edmd-optimization} and the encoder-decoder reconstruction loss
    \begin{equation*}
         \mathcal{L}_{\text{enc-dec}}(\vec{\theta}, \vec{\omega}) = \frac{1}{m} \sum_{i=1}^m \norm{\vec{x}^{(i)} - \vec{\varphi}_{\vec{\omega}} \left(\vec{\psi}_{\vec{\theta}}\big(\vec{x}^{(i)}\big)\right)}_2^2.
    \end{equation*}
Methods of this type are typically referred to as \emph{Koopman autoencoders} (KAEs). In practice, it has been observed that using multi-step forward prediction instead of the single-step loss in \eqref{eq:edmd-optimization} results in improved performance. Furthermore, it was demonstrated in Ref.~\onlinecite{AELM20} that additional performance gains can be made by learning matrices for both the forward and backward dynamics and enforcing that the product of the two is close to the identity (forward-backward consistency). The associated models are called \emph{consistent Koopman autoencoders} (cKAEs).

%%%%%%%%%%  EDMD WITH KERNEL DICTIONARY LEARNING
\section{EDMD with kernel-based dictionary learning}\label{sec:edmd-kdl}

We propose a novel approach, called \emph{extended dynamic mode decomposition with kernel-based dictionary learning} (EDMD-kDL), for learning finite-dimensional Koopman embeddings of nonlinear dynamics from data. In contrast to standard methods that rely on ANNs, EDMD-kDL optimizes EDMD dictionaries over an RKHS associated with a user-specified kernel. Furthermore, our approach is fundamentally different from standard kernel EDMD in that the dimensionality of learned dictionary is independent of the number of training data points. Additionally, EDMD-kDL allows the use of a multi-step prediction objection at training stage, in contrast to kernel EDMD.

\subsection{Derivation of EDMD-kDL}\label{sec:edmd-kdl-derivation}

Let $\H$ be the RKHS associated with a kernel $k\colon \Omega \times \Omega \to \R$. We are interested in learning a suitable $N$-valued dictionary map $\vec{\psi} \in \H^N$ for the data-driven approximation the Koopman operator. We require that the induced subspace contains, at least approximately, the full-state observable. To this end, we first focus on dictionaries of the form
\eqref{eq:dictionary-with-direct-link} and extend the approach to an autoencoder formulation in Section~\ref{sec:learned-decoder}. Furthermore, we consider training with a multi-step prediction objective in Section~\ref{sec:multistep-prediction}.

Assuming data of the form \eqref{eq:edmd-data-matrices}, we formulate the dictionary learning problem as solving
    \begin{equation}\label{eq:edmd-kdl}
       \left\{
            \begin{aligned}[c]
                &\underset{\substack{\vec{\phi} \in \H^{p} \\ K \in \R^{N\times N}}}{\min} \frac{1}{m} \sum_{i=1}^m \norm{\vec{\psi}\big(\vec{y}^{(i)}\big) - K \vec{\psi}\big(\vec{x}^{(i)}\big)}_2^2 + \beta \norm{\vec{\phi}}_{\H^{p}}^2\\
                &\text{s.t. }
             \vec{\psi}(\vec{x}) = \left[
                \vec{\rho}(\vec{x})^\top, \vec{\phi}(\vec{x})^\top\right]^\top,
            \end{aligned}
       \right.
    \end{equation}
where $\beta > 0$ is a regularization parameter. Unless the kernel $k$ admits a finite-dimensional feature map, \eqref{eq:edmd-kdl} involves an optimization problem over an infinite-dimensional function space. However, if function values for $\vec{\phi}$ were available at a finite set of points, we could obtain an equivalent finite-dimensional formulation due the particular RKHS structure of $\H$, specifically the \emph{representer theorem}~\cite{kimeldorf1971some, scholkopf2001generalized}. To obtain a finite-dimensional problem in the absence of function values and without an explicit parametrization of the RKHS (this could be achieved approximately by truncating the Mercer expansion of the kernel~\cite{Schoe01}, for example), we consider \emph{pseudo-values} $\left\{\vec{u}^{(j)}\right\}_{j=1}^{m_c} \subset \R^p$ at a finite set of \emph{collocation points} $\left\{\vec{z}^{(j)}\right\}_{j=1}^{m_c} \subset \Omega$; i.e., we require $\vec{\phi}\big(\vec{z}^{(j)}\big) = \vec{u}^{(j)}$ for each $j$ and aim to automatically learn the optimal matrix $U = \left[ \vec{u}^{(1)}, \dots, \vec{u}^{(m_c)}\right]$ from data. We then reformulate \eqref{eq:edmd-kdl} as a bilevel optimization problem:
    \begin{widetext}
        \begin{equation}\label{eq:edmd-kdl-bilevel-optim}
       \left\{
            \begin{aligned}[c]
                & \widehat{K}, \widehat{U} = \underset{\substack{K \in \R^{N\times N} \\ U \in \R^{p \times m_c}}}{\argmin} \: \frac{1}{m} \sum_{i=1}^m \norm{\vec{\widehat{\psi}}_U\big(\vec{y}^{(i)}\big) - K \vec{\widehat{\psi}}_U\big(\vec{x}^{(i)}\big)}_2^2 + \beta \norm{\vec{\widehat{\phi}}_U}_{\H^{p}}^2 &&& \text{(outer optimization)}\\
                &\text{s.t. } \quad \left\{
                \begin{aligned}
                    \vec{\widehat{\phi}}_U &=  \underset{\vec{\phi} \in \H^{p}}{\argmin} \ts \frac{1}{m_c} \sum_{j=1}^{m_c} \norm{\vec{\phi}\big(\vec{z}^{(j)}\big) - \vec{u}^{(j)}}_2^2 + \gamma  \norm{\vec{\phi}}_{\H^{p}}^2 \\
                    \vec{\widehat{\psi}}_U(\vec{x}) &= \left[
                \vec{\rho}(\vec{x})^\top, \vec{\widehat{\phi}}_U(\vec{x})^\top\right]^\top,
                \end{aligned}
                \right.
                 &&& \text{(inner optimization)}
            \end{aligned}
            \right.
        \end{equation}
    \end{widetext}
where $\gamma > 0$ is a \emph{nugget} parameter which allows us to treat the pseudo-values as noisy observations. That is, for the inner optimization problem, we fix the matrix of pseudo-values $U$ and optimize over the complementary dictionary $\vec{\phi}$. We show in what follows that this problem can be solved explicitly and the minimizer is a parametric function of the pseudo-values. This reduces the dictionary learning problem to finding the optimal~$U$. We then jointly solve for the optimal $U$ and the corresponding EDMD matrix $K$ in the outer optimization problem. A similar bilevel optimization framework was used in Ref.~\onlinecite{chen2021solving} for solving nonlinear partial differential equations with Gaussian processes. We provide more details below.

\subsubsection{Solving the inner optimization problem}\label{sec:edmd-kdl-first-level}

We start by observing that the $p$ components in the inner problem are decoupled, i.e., it can be rewritten as
    \begin{equation}\label{eq:inner-optim-decoupled}
       \widehat{\phi}_{U; \ell} =  \underset{\phi_{\ell} \in \H}{\argmin} \frac{1}{m_c} \sum_{j=1}^{m_c} \left(\phi_{\ell}\big(\vec{z}^{(j)}\big) - u_{\ell}^{(j)}\right)^2 + \gamma \norm{\phi_{\ell}}_{\H}^2,
    \end{equation}
 for $\ell = 1, \dots, p$, with $\phi_{\ell}$ and $u_{\ell}^{(j)}$ denoting the $\ell^{th}$ components of $\vec{\phi}$ and $\vec{u}^{(j)}$, respectively. The key ingredient for solving \eqref{eq:inner-optim-decoupled} is the following result.
    \begin{theorem}[Representer theorem~\cite{kimeldorf1971some, scholkopf2001generalized, Schoe01}]\label{thm:representer-theorem}
        Let $\mathcal{R}\colon [0, \infty) \to \R$ denote a strictly monotonically increasing function, $\mathbb{X}$ a set, and $\mathcal{L}\colon \left(\mathbb{X} \times \R^2\right)^m \to \R \cup \{\infty\}$ an arbitrary loss function. Then each minimizer $f \in \H$ of the regularized risk
            \begin{equation*}
                 \mathcal{L}\big(\big(x^{(i)}, y^{(i)}, f\big(x^{(i)}\big) \big)_{i=1}^m\big) + \mathcal{R}\left(\norm{f}_{\H}\right)
            \end{equation*}
        admits a representation of the form
            \begin{equation*}
                f(x) = \sum_{i=1}^m \alpha_i k\big(x^{(i)}, x\big).
            \end{equation*}
    \end{theorem}

 It then follows from this representer theorem that any solution of \eqref{eq:inner-optim-decoupled} must lie in the span of the kernel evaluated at the collocation points. That is, it is sufficient to consider functions of the form
    \begin{equation}\label{eq:edmd-kdl-representer-theorem}
        \phi_{\ell}(\ts\cdot\ts) = \sum_{j=1}^{m_c} q_{j\ell} k\big(\vec{z}^{(j)}, \ts\cdot\ts\big) = \vec{q}_{\ell}^\top \vec{k}_Z(\ts\cdot\ts),
    \end{equation}
where $\vec{q}_{\ell} \in \R^{m_c}$ is a vector of coefficients.
Given this representation, it is easy to show that $\big\|\phi_{\ell}\big\|_{\H}^2 = \vec{q}_{\ell}^\top G \ts \vec{q}_{\ell}$, where $G = G_{ZZ}$ is the Gram matrix evaluated at the collocation points. Therefore, \eqref{eq:inner-optim-decoupled} is equivalent to
    \begin{equation*}
       \vec{\widehat{q}}_{\ell}  = \underset{\vec{q}_{\ell} \in \R^{m_c}}{\argmin}\ts \overbrace{\frac{1}{m_c} \sum_{j=1}^{m_c} \left(\vec{q}_{\ell}^\top \vec{k}_Z\big(\vec{z}^{(j)}\big) - u_{\ell}^{(j)}\right)^2 + \gamma \vec{q}_{\ell}^\top G \vec{q}_{\ell}}^{\mathcal{J} \left(\vec{q}_{\ell}\right)}.
    \end{equation*}
The objective function can be written compactly as
    \begin{equation*}
        \mathcal{J}(\vec{q}_{\ell}) = \frac{1}{m_c} \Bigg\|
        \overbrace{
        \begin{bmatrix}
            \vec{k}_Z^\top\big(\vec{z}^{(1)}\big) \\
            \vdots \\
            \vec{k}_Z^\top\big(\vec{z}^{(m)}\big)
        \end{bmatrix}
        }^{G} \vec{q}_{\ell} -
            \overbrace{
            \begin{bmatrix}
                u_{\ell}^{(1)} \\
                \vdots \\
                u_{\ell}^{(m)}
            \end{bmatrix}
            }^{\vec{u}_{\ell}}
        \Bigg\|_2^2 + \gamma \vec{q}_{\ell}^\top G \vec{q}_{\ell},
    \end{equation*}
and its gradient with respect to $\vec{q}_{\ell}$ is given by
    \begin{equation*}
        \nabla_{\vec{q}_{\ell}}\mathcal{J} = \frac{2}{m_c} G \left(G \vec{q}_{\ell} - \vec{u}_{\ell}\right) + 2 \ts \gamma \ts G \ts \vec{q}_{\ell}.
    \end{equation*}
It follows from the first-order condition $\nabla_{\vec{q}_{\ell}}\mathcal{J}\left(\vec{\widehat{q}}_{\ell}\right) = \vec{0}$ that
    \begin{equation}\label{eq:inner-opt-first-order-cond}
        G\left[\left(G + \gamma m_c \ts \ts \mathrm{I}\right) \vec{\widehat{q}}_{\ell} - \vec{u}_{\ell}\right] = \vec{0},
    \end{equation}
which admits a unique solution given by
    \begin{equation}\label{eq:inner-opt-sol}
        \vec{\widehat{q}}_{\ell} = \left(G + \gamma \ts m_c \ts \mathrm{I}\right)^{-1} \vec{u}_{\ell},
    \end{equation}
provided that $G$ is invertible, where $\mathrm{I}$ denotes the identity matrix of dimension $m_c$. Note that even when $G$ is singular, the solution to \eqref{eq:inner-optim-decoupled} is still unique and given by
    \begin{equation}
        \widehat{\phi}_{U;\ell}(\ts\cdot\ts) = \vec{\widehat{q}}_{\ell}^\top \vec{k}_Z(\ts\cdot\ts),
    \end{equation}
despite \eqref{eq:inner-opt-first-order-cond} not having a unique solution.

\begin{remark}
It is worth noting that although we have provided detailed derivations for the sake of completeness and to increase accessibility for readers who are not familiar with kernel methods, we could have instead relied on the fact that \eqref{eq:inner-optim-decoupled} corresponds to a standard \emph{kernel ridge regression} problem~\cite{hoerl1970ridge}.
\end{remark}

\subsubsection{Solving the outer optimization problem} \label{sec:edmd-kdl-second-level}

We now turn our focus to solving the outer optimization problem in \eqref{eq:edmd-kdl-bilevel-optim}. To this end, we first calculate the squared RKHS norm of $\vec{\widehat{\phi}}_U$ as
    \begin{equation}\label{eq:squared-rkhs-norm}
        \begin{split}
            \Big\|\vec{\widehat{\phi}}_U\Big\|_{\H^p}^2 &= \sum_{\ell=1}^p \Big\|\widehat{\phi}_{U; \ell}\Big\|_{\H}^2 = \sum_{\ell=1}^p \vec{\widehat{q}}_{\ell}^\top G \ts \vec{\widehat{q}}_{\ell}\\
            &= \sum_{\ell=1}^p \vec{u}_{\ell}^\top \left(G + \gamma m_c \mathrm{I}\right)^{-1} G \left(G + \gamma m_c \mathrm{I}\right)^{-1} \vec{u}_{\ell}.
        \end{split}
    \end{equation}
This expression can be written more compactly. To see that, let $G = V \Sigma V^\top$ be the (compact) singular value decomposition (SVD) of the Gram matrix $G$. Note that the left and right singular vectors coincide here since $G$ is symmetric and positive semi-definite. Define $L = V\Sigma^{\nicefrac{1}{2}}$ and $Q = L^\top \left(G + \gamma m_c \mathrm{I}\right)^{-1}$.  Then $L L^\top = G$ and $Q^\top Q = \left(G + \gamma m_c \mathrm{I}\right)^{-1} G \left(G + \gamma m_c \mathrm{I}\right)^{-1}$. It follows that \eqref{eq:squared-rkhs-norm} is equivalent to
    \begin{equation*}
        \Big\|\vec{\widehat{\phi}}_U\Big\|_{\H^p}^2 = \tr\left[(QU^\top)^\top (Q U^\top)\right] = \norm{Q U^\top}_F^2,
    \end{equation*}
where $\norm{\ts\cdot\ts}_F$ denotes the Frobenius norm of a matrix.

Furthermore, we note that the variables $U$ and $K$ in the outer optimization problem are \emph{separable} in the sense that for fixed $U$, we can solve for $K$ exactly as in standard EDMD. Therefore, we can either directly optimize over both $U$ and $K$ jointly or use the \emph{variable projection method}~\cite{golub1973differentiation, golub2003separable} to obtain
    \begin{equation}\label{eq:edmd-kdl-outer-reduced}
        \left\{
            \begin{aligned}
                \widehat{U} &=  \underset{U\in \mathbb{R}^{p \times m_c}}{\argmin \ts} \frac{1}{m}
               \norm{\widehat{\Psi}_{U, Y} - \widehat{K}_U \widehat{\Psi}_{U, X} }_F^2 + \beta \norm{Q U^\top}_F^2,\\
               & \text{ s.t. }
               \left\{
               \begin{aligned}
                   \widehat{K}_U &= \underset{K \in \mathbb{R}^{(n+p) \times (n+p)}}{\argmin} \frac{1}{m} \norm{\widehat{\Psi}_{U, Y} - K \widehat{\Psi}_{U, X} }_F^2 \\
                    &= \widehat{\Psi}_{U, Y} \widehat{\Psi}_{U, X}^\top \left( \widehat{\Psi}_{U, X} \widehat{\Psi}_{U, X}^\top\right)^\dagger,
               \end{aligned}
               \right.
            \end{aligned}
        \right.
    \end{equation}
as the final optimization problem over $U$. We use the L-BFGS optimizer~\cite{liu1989limited}, a quasi-Newton method, with automatic differentiation as implemented in the PyTorch library to solve \eqref{eq:edmd-kdl-outer-reduced}. Alternatively, one could calculate an analytic expression for the gradient of the objective function and use standard numerical optimization routines. We summarize the approach in Algorithm~\ref{alg:edmd-kdl}.

 \begin{widetext}
    \SetKwComment{Comment}{/* }{ */}
    \begin{algorithm}
        \caption{EDMD-kDL algorithm}\label{alg:edmd-kdl}
        \KwData{kernel $k$, data matrices $X, Y \in \R^{n\times m}$, collocation points $Z \in \R^{n \times m_c}$, number of learnable dictionary elements $p \geq 1$, fixed component $\vec{\rho} \colon \Omega \to \R^o$, and regularization parameters $\gamma, \beta > 0$}
        \KwResult{learned dictionary $\vec{\widehat{\psi}}$, EDMD matrix $\widehat{K}$, and approximated Koopman spectral information}
        Assemble the Gram matrix $G = k(Z, Z)$\;
        Perform a (compact) SVD of $G$; i.e., $G = V\Sigma V^\top$ \;
        $L \gets V\Sigma^{\nicefrac{1}{2}} $\;
        $Q \gets L^T\left(G + \gamma m_c \mathrm{I}\right)^{-1}$\;
        Solve \eqref{eq:edmd-kdl-outer-reduced} for the optimal $\widehat{U} = \begin{bmatrix}
            \vec{\widehat{u}}_1^\top \\ \vdots \\ \vec{\widehat{u}}_p^\top
        \end{bmatrix}$ \Comment*[r]{using L-BFGS or your preferred optimizer}
        \For{$\ell = 1, \dots, p$}{
          $\vec{\widehat{q}}_{\ell} \gets \left(G + \gamma m_c \mathrm{I}\right)^{-1} \vec{\widehat{u}}_{\ell}$ \Comment*[r]{learned dictionary elements}
          $\widehat{\phi}_{\widehat{U}; \ell} \gets \vec{\widehat{q}}_{\ell}^\top \vec{k}_Z(\ts\cdot\ts) $
        }
        $\vec{\widehat{\phi}}_{\widehat{U}} \gets \left[\widehat{\phi}_{\widehat{U}; 1}, \dots, \widehat{\phi}_{\widehat{U}; p}\right]^\top $\;
        $\vec{\widehat{\psi}} \gets \left[\vec{\rho}^\top,\:  \vec{\widehat{\phi}}_{\widehat{U}} ^\top\right]^\top$  \Comment*[r]{assemble full dictionary}
        $\widehat{K} \gets \widehat{\Psi}_Y \widehat{\Psi}_X^\dagger$\Comment*[r]{EDMD matrix}
        Compute eigenvalues, eigenfunctions, and modes as in standard EDMD
    \end{algorithm}
\end{widetext}

\subsection{EDMD-kDL with kernel-based autoencoders}\label{sec:learned-decoder}

We now extend our derivation to an autoencoder formulation where, instead of including fixed features, we simultaneously learn the full dictionary and a decoder map that is trained to reconstruct the full-state observable. We call this new type of architecture \emph{kernel-based autoencoders}.

Since one of the prominent applications of EDMD is the decomposition of the full-state observable into a linear combination of (approximate) Koopman eigenfunctions, which requires the full-state observable to be in the linear span of the dictionary, we constrain the decoder to be linear. Specifically, we formulate the new learning task as solving
    \begin{equation}\label{eq:edmd-kdl-decoder}
       \left\{
            \begin{aligned}[c]
                &\underset{\substack{\vec{\psi} \in \H^{N} \\ K \in \R^{N\times N} \\ D \in \R^{n\times N} }}{\argmin} \frac{1}{m} \sum_{i=1}^m \norm{\vec{\psi}\big(\vec{y}^{(i)}\big) - K \vec{\psi}\big(\vec{x}^{(i)}\big)}_2^2 + \beta \norm{\vec{\psi}}_{\H^{N}}^2\\
                &\text{s.t. }
             \vec{x} = D \vec{\psi}(\vec{x}), \quad \forall \vec{x} \in \Omega.
            \end{aligned}
       \right.
    \end{equation}
Then, following the same strategy as in Section \ref{sec:edmd-kdl-derivation}, we obtain a bilevel reformulation by introducing pseudo-values at a finite set of collocations points. The inner problem, with solution denoted by $\vec{\widehat{\psi}}_U$, is the same as in \eqref{eq:edmd-kdl-bilevel-optim}, with the difference that $\vec{\phi}$ and $\H^p$ are now replaced by $\vec{\psi}$ and $\H^N$, respectively. For the outer problem, the objective function is a linear combination of the following loss functions:
    \begin{enumerate}[leftmargin=3.5ex, itemsep=0ex, topsep=0.5ex, label=\roman*)]
        \item The standard EDMD loss
            \begin{equation}\label{eq:outer-level-edmd-loss}
               \mathcal{L}_1(K, D, U) =  \frac{1}{m} \sum_{i=1}^m \norm{\vec{\widehat{\psi}}_U\big(\vec{y}^{(i)}\big) - K \vec{\widehat{\psi}}_U\big(\vec{x}^{(i)}\big)}_2^2.
            \end{equation}
        \item The autoencoder reconstruction loss
            \begin{equation}\label{eq:outer-level-decoder-loss}
               \mathcal{L}_2(K, D, U) =  \frac{1}{m_c} \sum_{j=1}^{m_c} \norm{\vec{z}^{(j)} - D \ts \vec{\widehat{\psi}}_U\big(\vec{z}^{(j)}\big)}_2^2,
            \end{equation}
            which imposes the soft encoder--decoder constraint $\vec{x} = D \vec{\psi}(\vec{x})$ at the collocation points $\big\{\vec{z}^{(j)}\big\}_{j=1}^{m_c}$. Note that, in principle, we could choose a different set of points. For simplicity, however, we choose this.
        \item The RKHS regularization
            \begin{equation}\label{eq:outer-level-rkhs-regularization}
                \mathcal{L}_3(K, D, U) = \big\|\vec{\widehat{\psi}}_U\big\|_{\H^N}^2 = \norm{Q U^\top}_F^2.
            \end{equation}
    \end{enumerate}
It is clear that if $U$ is fixed, we can solve the optimization problems for obtaining $K$ and $D$ separately. Therefore, as in the previous section, we can either directly optimize over $K$, $D$, and $U$ simultaneously, or use variable projection to obtain
    \begin{equation}
       \widehat{U} = \underset{U\in \mathbb{R}^{N \times m_c}}{\argmin \ts}  \sum_{i=1}^3 \beta_i \ts \mathcal{L}_i\big(\widehat{K}_U, \widehat{D}_U, U \big),
    \end{equation}
where the $\beta_i$'s are positive weights.  Here, $\widehat{K}_U$ is as in \eqref{eq:edmd-kdl-outer-reduced} and  $\widehat{D}_U$ is given by
    \begin{equation}
        \begin{split}
            \widehat{D}_U &= \underset{D \in \mathbb{R}^{n \times N}}{\argmin}\, \frac{1}{m_c}\sum_{j=1}^{m_c} \norm{\vec{z}^{(j)} - D \vec{\widehat{\psi}}_U\big(\vec{z}^{(j)}\big) }_2^2 \\
            &= Z \widehat{\Psi}_{U, Z}^\top \left(\widehat{\Psi}_{U, Z} \widehat{\Psi}_{U, Z}^\top\right)^\dagger,
        \end{split}
    \end{equation}
with $Z = \left[\vec{z}^{(1)}, \dots, \vec{z}^{(m_c)} \right]$.

\subsection{Learning with a multi-step prediction objective}\label{sec:multistep-prediction}
Thus far, we have optimized EDMD dictionaries with a single-step prediction objective. However, it has been observed that training ANN-based Koopman autoencoders using multi-step prediction leads to higher accuracy and robustness~\cite{Otto19,AELM20}. We expect a similar pattern for the proposed framework and therefore extend it to this setting. Considering the kernel-based autoencoder formulation, we now assume we have access to training data of the form $\left\{\left( \vec{x}^{(\ell)}, Y^{(\ell)} \right)\right\}_{\ell = 1}^m$, where
    \begin{equation*}
        Y^{(\ell)} = \left[\vec{y}^{(\ell)}_1,\, \vec{y}^{(\ell)}_2,\, \dots,\, \vec{y}^{(\ell)}_T \right],
    \end{equation*}
with $\vec{y}^{(\ell)}_t = \vec{f}^t\big(\vec{x}^{(\ell)}\big)$, for $t = 1, \dots, T$. Here, $T \geq 1$ is the number of forward prediction steps. Then, using pseudo-values at a set of collocation points to obtain a bilevel formulation again, the final optimization problem can be written as
    \begin{equation*}
        \widehat{K}, \widehat{D}, \widehat{U} = \underset{K, D, U}{\arg\min} \sum_{i=1}^3 \beta_i \ts \mathcal{L}_i(K,D,U),
    \end{equation*}
where
    \begin{equation*}
        \mathcal{L}_1 = \frac{1}{m} \sum_{\ell=1}^m \frac{1}{T} \sum_{t=1}^T \norm{\vec{\widehat{\psi}}_U\big(\vec{y}^{(\ell)}_t\big) - K^t \vec{\widehat{\psi}}_U\big(\vec{x}^{(\ell)}\big) }_2^2,
    \end{equation*}
and $\mathcal{L}_2$ and $\mathcal{L}_3$ are as in \eqref{eq:outer-level-decoder-loss} and \eqref{eq:outer-level-rkhs-regularization}, respectively. Additionally, it is possible to use variable projection for the decoder matrix $D$ as in Section~\ref{sec:learned-decoder}, but not for the Koopman matrix $K$.

\section{Numerical experiments}\label{sec:numerical-experiments}

In this section, we illustrate the performance of EDMD-kDL on a wide range of benchmark problems. We consider synthetic data from the model of an undamped nonlinear pendulum and the simulation of a fluid flow past a cylinder exhibiting vortex shedding. Additionally, we consider real-world datasets. We test the performance of EDMD-kDL on the task of forecasting global sea-surface temperature (SST) and predicting the next frame of a video of a pendulum. We compare the proposed kernel-based methods against state-of-the art ANN-based alternatives and EDMD with random Fourier features (EDMD-RFF)~\cite{degennaro2019scalable, NK23}, which is an approximation of the standard kernel EDMD method. In all experiments, collocation points for EDMD-kDL are generated using a Sobol sequence~\cite{sobol1967distribution} and we use a Gaussian RBF kernel with bandwidth $\sigma$. In addition, we use the official implementations of the ANN-based approaches proposed in Ref.~\onlinecite{AELM20}, publicly available at \href{https://github.com/erichson/koopmanAE}{github.com/erichson/koopmanAE}.

\subsection{Undamped nonlinear pendulum}

The dynamics of a frictionless pendulum with no forcing is modeled by the two-dimensional Hamiltonian system
    \begin{equation*}
        \left\{\begin{aligned}
            \dot{x}_1 &= x_2, \\
            \dot{x}_2 &= -\frac{g}{\ell} \sin{x_1},
        \end{aligned}\right.
    \end{equation*}
where $x_1$ (in radians) is the angular position, $x_2 \,(m/s)$  is the velocity, $g = 9.81 \,(m/s^2)$ is the gravitational constant, and $\ell = 1\, (m)$ is the string's length. We assume throughout that $x_2(0) = 0$. It follows that the degree of nonlinearity of the dynamics is determined by the value of the initial angular position $x_1(0)$.

Following Ref.~\onlinecite{AELM20}, we generate $1600$ data points from a single trajectory simulated with time step $\Delta t = 0.1$ using the exact solution given in terms of elliptical functions and consider two scenarios: an almost-linear regime with $x_1(0) = 0.8$ and a truly nonlinear regime with $x_1(0) = 2.4$. We use data from the first $600$ time steps for training and the remaining $1000$ steps for testing.

Throughout all of the experiments, we set the embedding dimension to $N=6$ and the number of forward steps to $T = 8$ for EDMD-kDL and Koopman autoencoders (KAE and cKAE). For the consistent KAE approach, the number of backward steps is also set to $8$. The bandwidth of the kernel is set to $\sigma = 0.75$ for EDMD-kDL and EDMD-RFF. Furthermore, we use this simple model to investigate the impact of noise on the performance of each method.

\subsubsection{Clean data setting}
\begin{figure*}
        \begin{subfigure}{.55\textwidth}
            \centering
            \caption{Predicted phase portraits, $x_1(0) = 0.8$.}
            \includegraphics[width=0.9\linewidth]{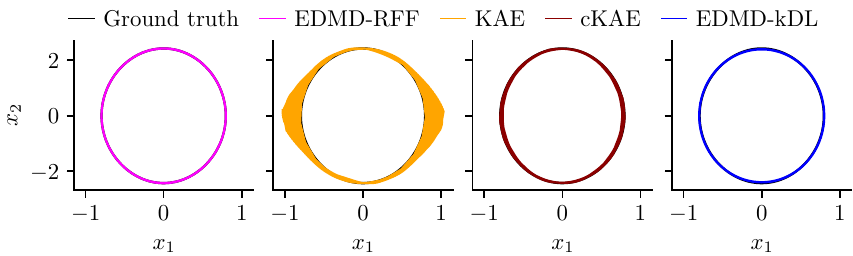}
            \label{fig:pendulum-noise-free-a}
        \end{subfigure}
        \begin{subfigure}{.4\textwidth}
            \centering
            \caption{Relative prediction error, $x_1(0) = 0.8$.}
            \includegraphics[width=0.9\linewidth]{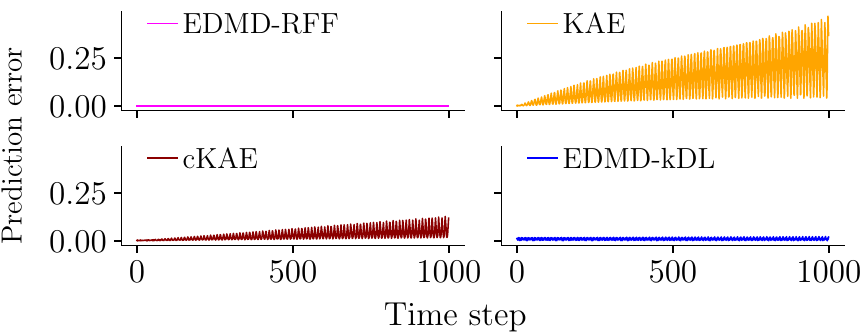}
            \label{fig:pendulum-noise-free-b}
        \end{subfigure}
        \newline
        \begin{subfigure}{.55\textwidth}
            \centering
            \caption{Predicted phase portraits, $x_1(0) = 2.4$.}
            \includegraphics[width=0.9\linewidth]{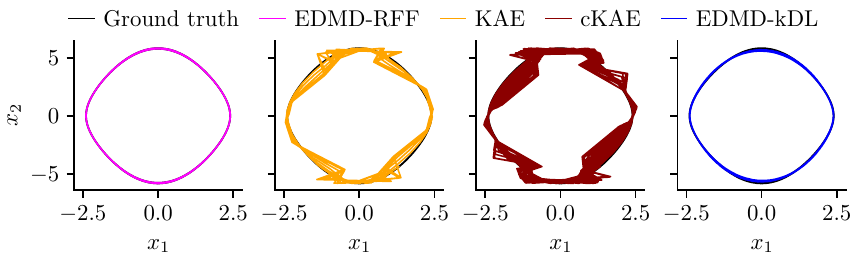}
            \label{fig:pendulum-noise-free-c}
        \end{subfigure}
        \begin{subfigure}{.4\textwidth}
            \centering
            \caption{Relative prediction error, $x_1(0) = 2.4$.}
            \includegraphics[width=0.9\linewidth]{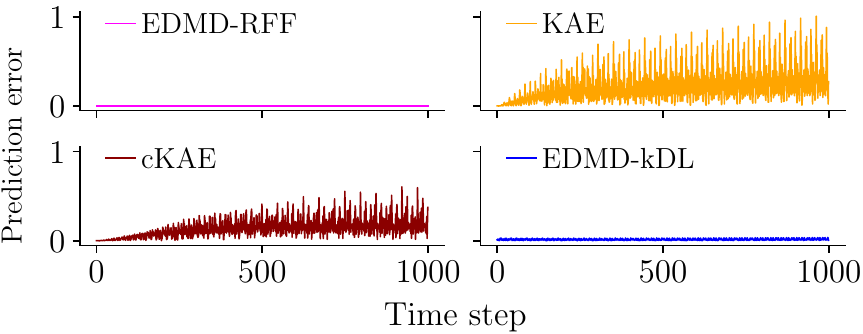}
            \label{fig:pendulum-noise-free-d}
        \end{subfigure}
        \caption{\textbf{Nonlinear pendulum in the clean data setting.} Phase portraits and relative prediction errors over $1000$ time steps of testing using learned Koopman models. The top and bottom rows correspond to the almost-linear and truly nonlinear regimes, respectively.}
        \label{fig:pendulum-noise-free}
    \end{figure*}

We start with the noise-free data scenario. Figure~\ref{fig:pendulum-noise-free} shows the phase portraits and prediction errors over the testing horizon. For each method, predictions are generated autoregressively by simulating the learned linear Koopman models, using the first testing point as initial condition, and then mapping to the state space using the learned decoders. We observe that for this simple problem, EDMD-RFF achieves the lowest prediction error, suggesting that the RFF dictionary -- which consists of sinusoidal functions -- is well-suited for the problem. However, among the dictionary learning frameworks, EDMD-kDL consistently outperforms the neural network approaches. Furthermore, despite cKAE performing clearly better than standard KAE in accordance with observations made in Ref.~\onlinecite{AELM20}, they both fail to properly capture the underlying phase portrait of the system in the truly nonlinear regime (Figures~\ref{fig:pendulum-noise-free-c} \& \ref{fig:pendulum-noise-free-d}).

\subsubsection{Noisy data setting}
To investigate the robustness of the proposed method with respect to noise, we corrupt the pendulum data with mean-free Gaussian noise. To account for varying scales between the two sets of initial conditions, the noise scale $\sigma_{\text{noise}}$ is set according to
    \begin{equation*}
        \sigma_{\text{noise}} = \text{NR} \times \frac{\norm{X_{\text{traj}}}_F}{\sqrt{n \times M}},
    \end{equation*}
where $\text{NR} \in (0, 1)$ denotes the noise ratio and $X_{\text{traj}} \in \R^{n\times M}$ denotes the complete trajectory data with $M = 1600$. We experiment with noise ratios going from $1\%$ to $10\%$, specifically $\text{NR} \in \{0.01, 0.05, 0.1\}$. For the sake of fair comparison, we keep all hyperparameters fixed to the same values as in the clean data case for all methods.

\paragraph{The almost-linear regime.}

\begin{figure*}
        %%%%%%%% noise-level 1 %%%%%%%%%%%%%%
        \begin{subfigure}{.55\textwidth}
            \centering
            \caption{Predicted phase portraits, $\text{NR} = 1\%$.}
            \includegraphics[width=0.9\linewidth]{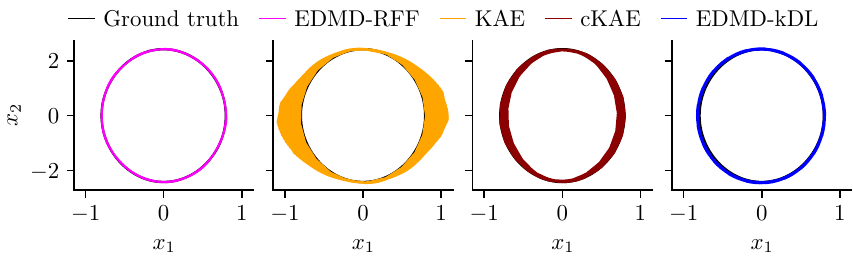}
            %\label{fig:pendulum-noise-free-a}
        \end{subfigure}
        \begin{subfigure}{.4\textwidth}
            \centering
            \caption{Relative prediction error, $\text{NR} =1\%$.}
            \includegraphics[width=0.9\linewidth]{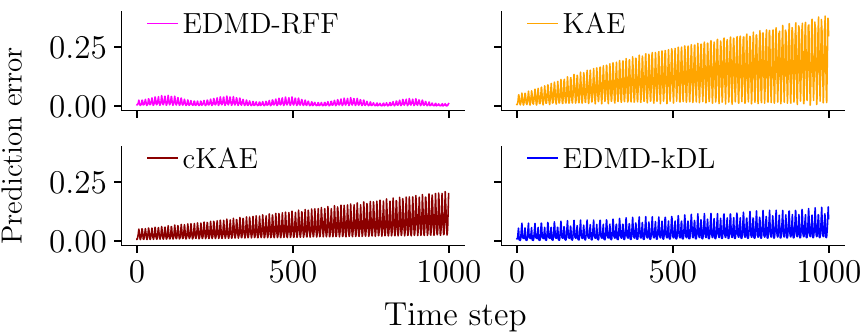}
            %\label{fig:pendulum-noise-free-a}
        \end{subfigure}
        \newline
        %%%%%%%% noise-level 2 %%%%%%%%%%%%%%
        \begin{subfigure}{.55\textwidth}
            \centering
            \caption{Predicted phase portraits, $\text{NR} =5\%$.}
            \includegraphics[width=0.9\linewidth]{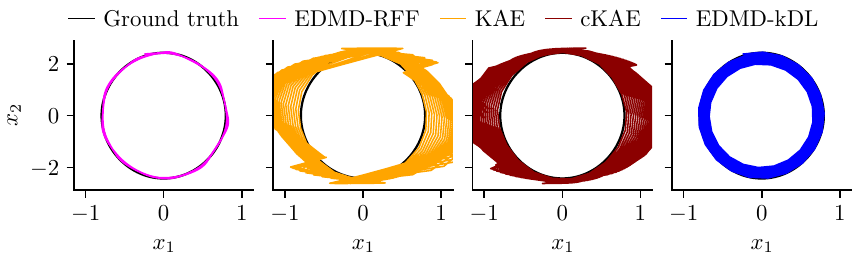}
            %\label{fig:pendulum-noise-free-b}
        \end{subfigure}
        \begin{subfigure}{.4\textwidth}
            \centering
            \caption{Relative prediction error, $\text{NR} =5\%$.}
            \includegraphics[width=0.9\linewidth, height=0.4\linewidth]{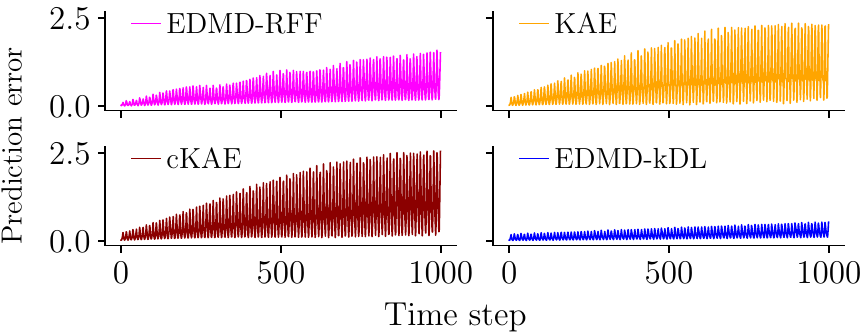}
            %\label{fig:pendulum-noise-free-b}
        \end{subfigure}
        \newline
        %%%%%%%% noise level 3 %%%%%%%%%%%%%%
        \begin{subfigure}{.55\textwidth}
            \centering
            \caption{Predicted phase portraits, $\text{NR} =10\%$.}
            \includegraphics[width=0.9\linewidth]{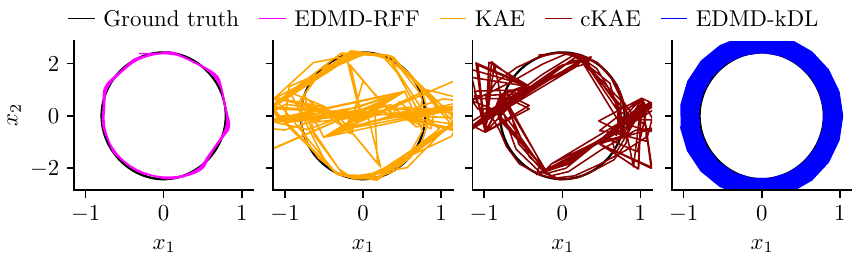}
            %\label{fig:pendulum-noise-free-c}
        \end{subfigure}
        \begin{subfigure}{.4\textwidth}
            \centering
            \caption{Relative prediction error, $\text{NR} = 10\%$.}
            \includegraphics[width=0.9\linewidth]{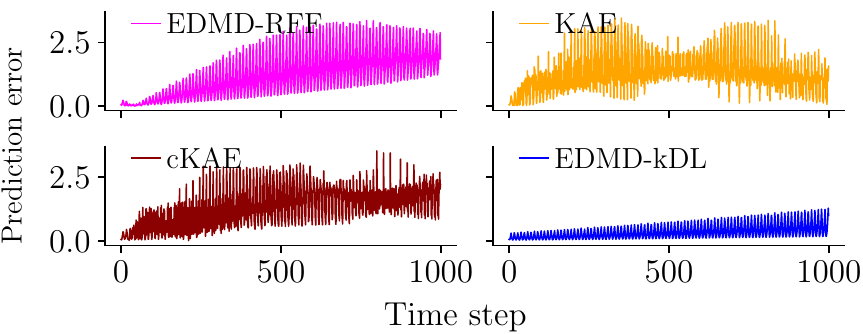}
            %\label{fig:pendulum-noise-free-c}
        \end{subfigure}
        \caption{\textbf{Pendulum with measurement noise in the almost-linear regime.} Across noise levels, EDMD-kDL outperforms the two Koopman autoencoder methods, which produce non-smooth predictions. On the other hand, EDMD-RFF suffers from phase drift (see the left column of Figure~\ref{fig:pendulum-noisy-trajectories}) which leads to worse performance in terms of prediction error compared to EDMD-kDL when the amount of noise is large.}
        \label{fig:pendulum-noisy-almost-linear}
    \end{figure*}

\begin{figure*}
        %%%%%%%% noise-level 2 %%%%%%%%%%%%%%
        \begin{subfigure}{.48\textwidth}
            \centering
            \caption{$x_1(0) = 0.8$ and $\text{NR} =5\%$.}
            \includegraphics[width=0.9\linewidth]{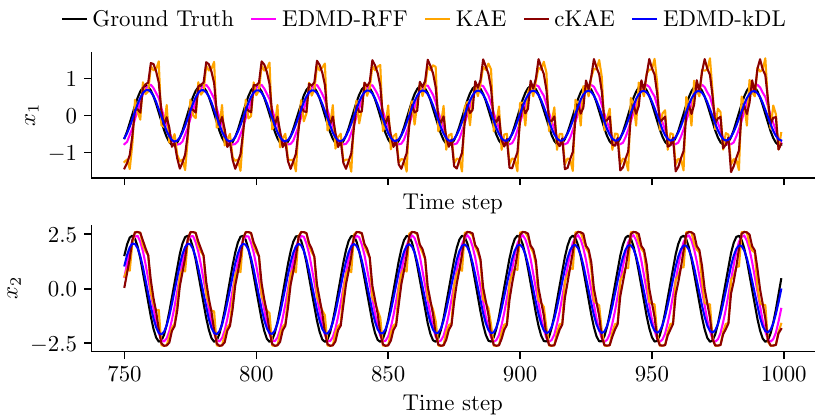}
            \label{fig:pendulum-noisy-trajectories-c}
        \end{subfigure}
        \begin{subfigure}{.48\textwidth}
            \centering
            \caption{$x_1(0) = 2.4$ and $\text{NR} =5\%$.}
            \includegraphics[width=0.9\linewidth]{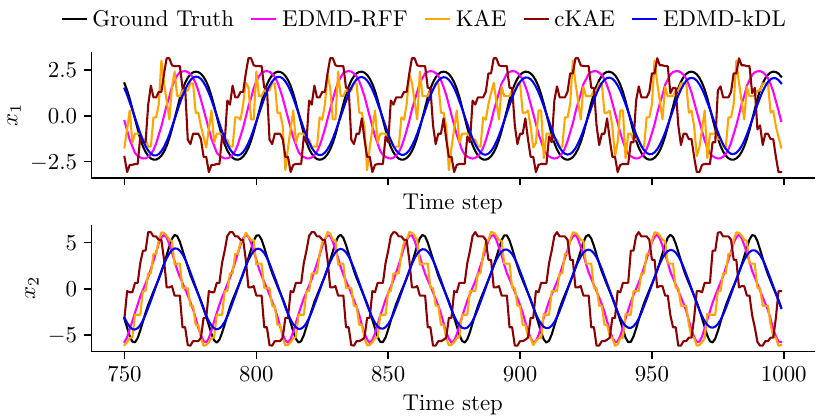}
            \label{fig:pendulum-noisy-trajectories-d}
        \end{subfigure}
        \newline
        %%%%%%%% noise-level 3 %%%%%%%%%%%%%%
        \begin{subfigure}{.48\textwidth}
            \centering
            \caption{$x_1(0) = 0.8$ and $\text{NR} =10\%$.}
            \includegraphics[width=0.9\linewidth]{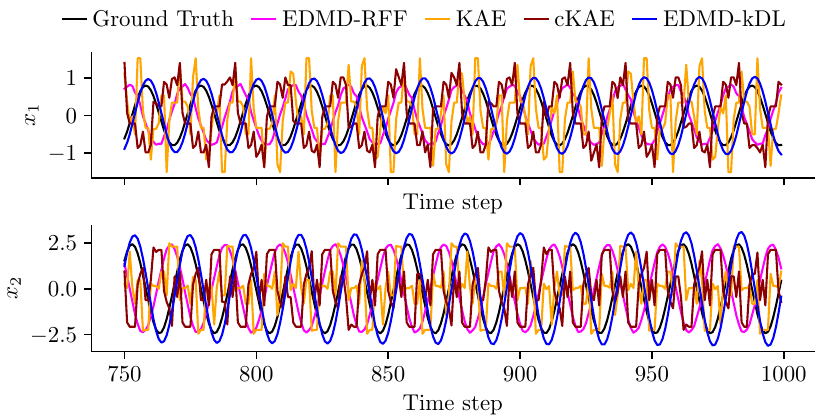}
            \label{fig:pendulum-noisy-trajectories-e}
        \end{subfigure}
        \begin{subfigure}{.48\textwidth}
            \centering
            \caption{$x_1(0) = 2.4$ and $\text{NR} =10\%$.}
            \includegraphics[width=0.9\linewidth]{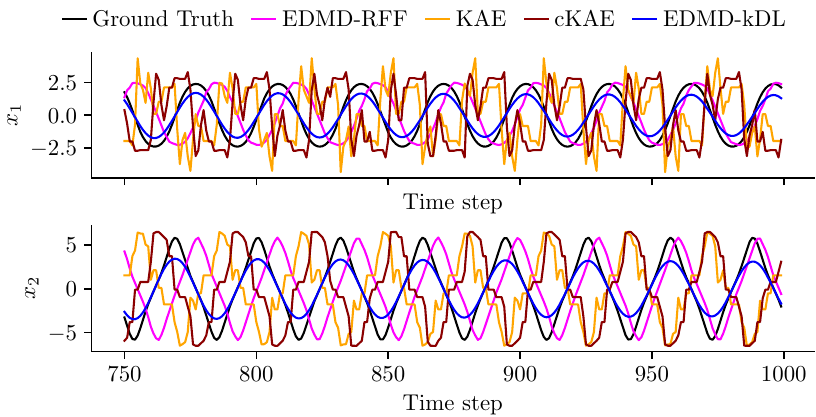}
            \label{fig:pendulum-noisy-trajectories-f}
        \end{subfigure}
        \caption{\textbf{Pendulum with measurement noise.} Comparison of predicted trajectories obtained by EDMD-kDL and three baseline methods over the final quarter of the prediction horizon. EDMD-RFF suffers from phase drift when the noise level is substantial whereas KAE and cKAE produce non-smooth predictions.}
        \label{fig:pendulum-noisy-trajectories}
    \end{figure*}

Figure~\ref{fig:pendulum-noisy-almost-linear} shows the phase portraits and prediction errors over time for each of the four methods. We observe that in the low-noise setting ($\text{NR} = 1\%$), EDMD-kDL performs marginally worse compared to EDMD-RFF as was the case in the clean data scenario. Interestingly, however, the relative prediction error incurred by EDMD-RFF grows substantially faster than that of EDMD-kDL when the noise level is high (second and third rows). To see why this happens despite the phase portraits suggesting the opposite, we zoom in at the final quarter of the predicted trajectories. Figures~\ref{fig:pendulum-noisy-trajectories-c} \& \ref{fig:pendulum-noisy-trajectories-e} show that despite properly preserving the amplitude, EDMD-RFF incurs a substantial phase drift from the ground truth solution, which explains the observed high error levels. In contrast, the model learned by EDMD-kDL produces a small error in the amplitude, but this is compensated by an accurate preservation of the phase compared to EDMD-RFF, which then results in lower errors overall. Furthermore, across noise levels, EDMD-kDL performs substantially better than KAE and cKAE in terms of both the prediction error and and the phase portrait preservation.

\begin{figure*}
        %%%%%%%% noise-level 1 %%%%%%%%%%%%%%
        \begin{subfigure}{.55\textwidth}
            \centering
            \caption{Predicted phase portraits, $\text{NR} = 1\%$.}
            \includegraphics[width=0.9\linewidth]{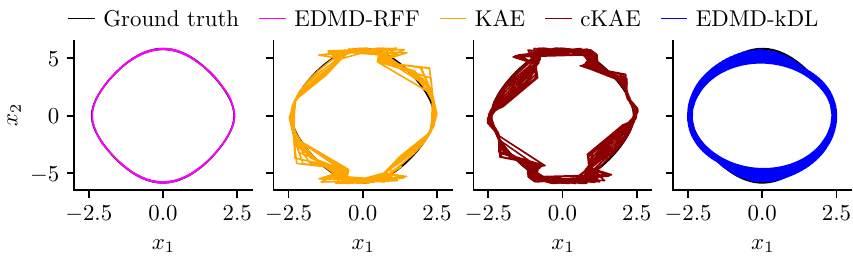}
            %\label{fig:pendulum-noise-free-a}
        \end{subfigure}
        \begin{subfigure}{.4\textwidth}
            \centering
            \caption{Relative prediction error, $\text{NR} =1\%$.}
            \includegraphics[width=0.9\linewidth]{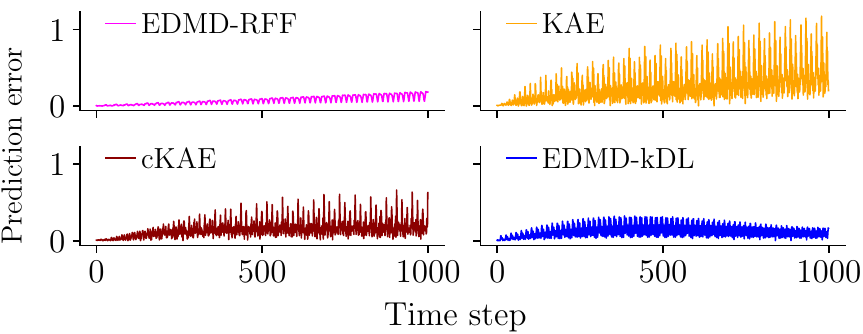}
            %\label{fig:pendulum-noise-free-a}
        \end{subfigure}
        \newline
        %%%%%%%% noise-level 2 %%%%%%%%%%%%%%
        \begin{subfigure}{.55\textwidth}
            \centering
            \caption{Predicted phase portraits, $\text{NR} =5\%$.}
            \includegraphics[width=0.9\linewidth]{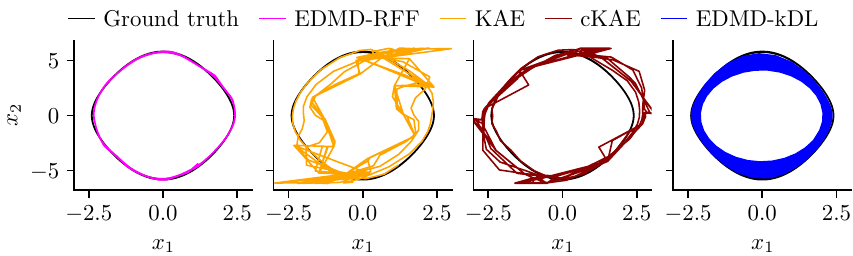}
            %\label{fig:pendulum-noise-free-b}
        \end{subfigure}
        \begin{subfigure}{.4\textwidth}
            \centering
            \caption{Relative prediction error, $\text{NR} =5\%$.}
            \includegraphics[width=0.9\linewidth]{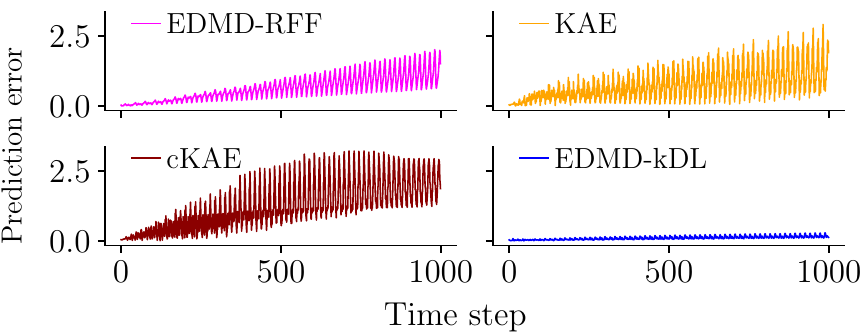}
            %\label{fig:pendulum-noise-free-b}
        \end{subfigure}
        \newline
        %%%%%%%% noise level 3 %%%%%%%%%%%%%%
        \begin{subfigure}{.55\textwidth}
            \centering
            \caption{Predicted phase portraits, $\text{NR} =10\%$.}
            \includegraphics[width=0.9\linewidth]{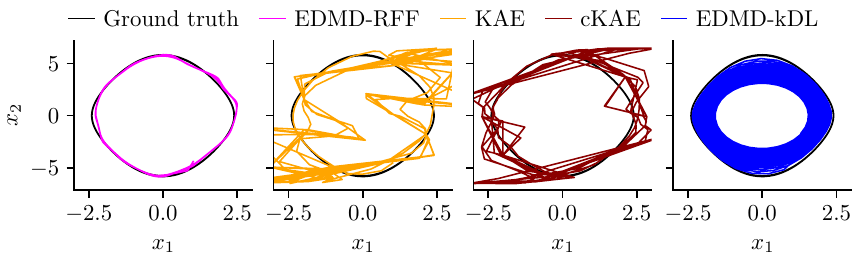}
            %\label{fig:pendulum-noise-free-c}
        \end{subfigure}
        \begin{subfigure}{.4\textwidth}
            \centering
            \caption{Relative prediction error, $\text{NR} = 10\%$.}
            \includegraphics[width=0.9\linewidth]{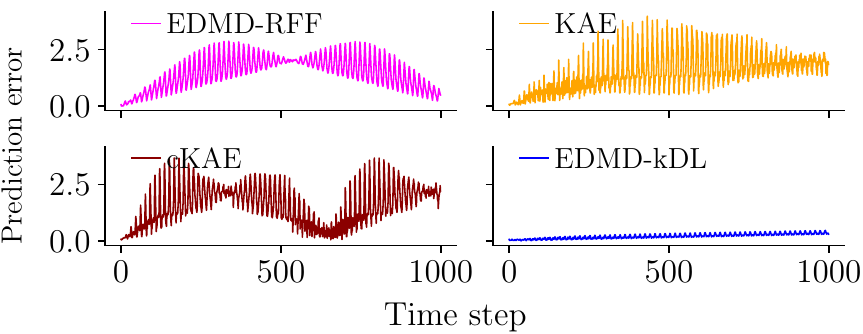}
            %\label{fig:pendulum-noise-free-c}
        \end{subfigure}
        \caption{\textbf{Pendulum with measurement noise in the truly nonlinear regime.} Across noise levels, EDMD-kDL outperforms the two Koopman autoencoder methods, which produce non-smooth predictions. On the other hand, EDMD-RFF suffers from phase drift (see the right column of Figure~\ref{fig:pendulum-noisy-trajectories}), which leads to worse performance in terms of prediction error compared to EDMD-kDL when the amount of noise is substantial.}
        \label{fig:pendulum-noisy-truly-nonlinear}
    \end{figure*}

\paragraph{The truly nonlinear regime.}
Figure~\ref{fig:pendulum-noisy-truly-nonlinear} shows the phase portraits and prediction errors over time for the four methods across the three noise levels. Additionally, the right panel of Figure~\ref{fig:pendulum-noisy-trajectories} shows the corresponding trajectories over the final quarter of the prediction horizon. The observations are similar to the almost-linear setting: EDMD-RFF best preserves the amplitude of the system. However, when the noise level is considerable, it suffers from a phase drift that results in high prediction errors. In contrast, EDMD-kDL properly balances the preservation of phase and amplitude and hence outperforms all the other methods in terms of prediction errors when the noise level is substantial. Moreover, Figure~\ref{fig:pendulum-noisy-trajectories} shows that models learned by the ANN-based approaches produce predictions that are highly irregular even when they are trained with forward-backward consistency.
    %%%%%% ALMOST-LINEAR REGIME %%%%%%%

    %%%%%% TRULY NONLINEAR REGIME %%%%%%%

    %%%%%% PREDICTED TRAJECTORIES %%%%%%%%%%%%%%

%%%%%%%%%%%%%%%%%%%%%%%%
\subsection{K\'{a}rm\'{a}n vortex shedding}

As a second classical benchmark problem, we consider fluid flow past a cylinder in two dimensions and choose the Reynolds number $ \mathrm{Re} = 100 $. We discretize the domain using a regular grid of size $ 180 \times 520 $ and generate $ m = 500 $ snapshots. We then reshape the snapshots into $93\ts600$-dimensional column vectors and assemble them into a single matrix $S$ with columns ordered sequentially in time. In order to reduce the dimensionality, we perform an SVD of the mean-subtracted snapshot matrix; i.e., $\bar{S} = U \Sigma V^\top$, and project the data onto the dominant $r = 19$ principal components, retaining $95\%$ of the cumulative energy. Here, $\bar{S}$ denotes the matrix obtained by subtracting the column-wise mean of $S$ from each of its columns. We use the first $400$ snapshots for training and the remaining $100$ snapshots for testing. In the previous example, we used the auto-encoder formulation for EDMD-kDL. We now choose the full-state observable as a fixed dictionary component instead and learn a complementary part of dimension $p = 2r$. On the other hand, we find that KAE and cKAE perform best with a lower embedding dimension and hence set it to $N = 6$ after some manual hyperparameter tuning.

\begin{figure}
    \centering
    \includegraphics[width=0.7\linewidth]{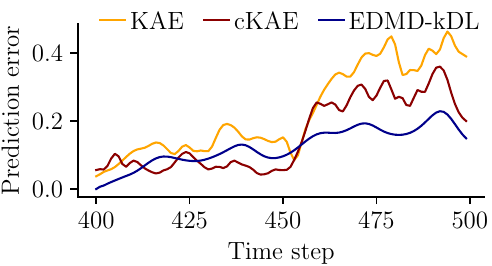}
    \caption{\textbf{K\'{a}rm\'{a}n vortex shedding}. Relative prediction errors over the testing horizon measured in the SVD-reduced space.}
    \label{fig:karman_vortex_error_plot}
\end{figure}

\begin{figure*}
    \centering
    \includegraphics[width=0.8\linewidth]{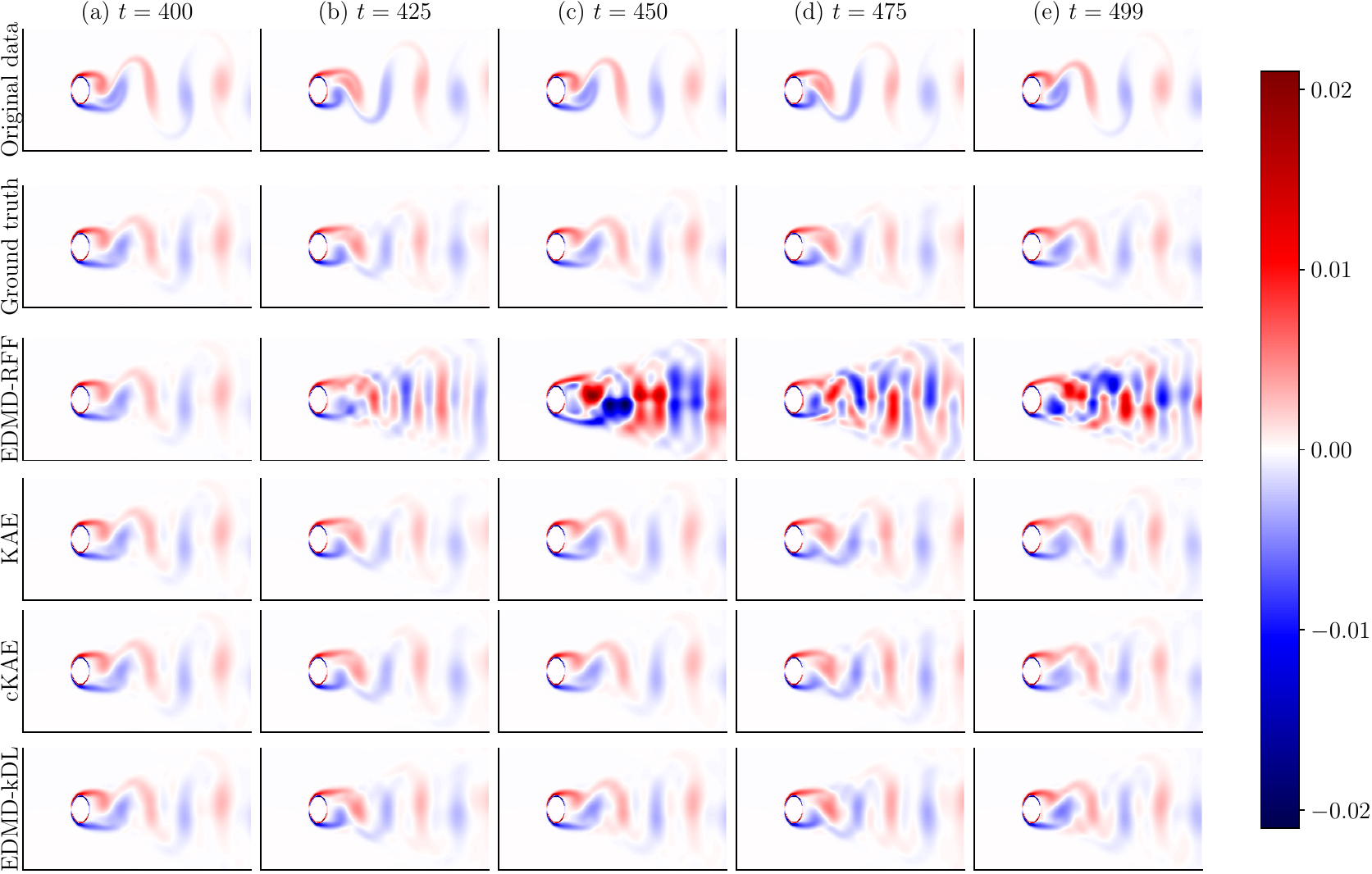}
    \caption{\textbf{K\'{a}rm\'{a}n vortex shedding}. Full-space predictions obtained by various Koopman methods. ``Ground truth'' refers to data obtained by projecting the original test data onto the retained principal components and reconstructing.}
    \label{fig:karman_vortex_snapshots}
\end{figure*}

Figure~\ref{fig:karman_vortex_error_plot} shows the obtained relative prediction errors over the testing horizon.  EDMD-RFF is not included in the plot since it incurred much higher errors, in contrast to the preceding example. The corresponding predicted snapshots are shown in Figure~\ref{fig:karman_vortex_snapshots}. We observe that all the methods except EDMD-RFF predict the dynamics of the system reasonably well. Of these, KAE performs the worst overall, whereas EDMD-kDL has the lowest error for long-term prediction.

%%%%%%%%%%%%%%%%%%%%%%%%
\subsection{Learning from video data}
Thus far, we have only considered experiments based on synthetic data. To illustrate the performance of the proposed method on real-world data, we consider the problem of predicting a moving pendulum directly from video data. The raw dataset, obtained from \href{http://www.youtube.com/watch?v=MpzaCCbX-z4}{ScienceOnline: The Pendulum and Galileo}, consists of a $20.92$ seconds video recorded at a $25$ frames per second rate on a $720 \times 576$ pixel grid. The authors of Ref.~\onlinecite{KSM19} used this dataset to illustrate the data-driven approximation of Koopman eigenfunctions using kernel EDMD. As a preprocessing step, we convert the video from RGB to grayscale with normalized values between zero and one and subtract the median frame to remove the background. This results in a dataset of size $720 \times 576 \times 523$, which we reshape into a single $414\ts720 \times 523$ snapshot matrix. We use the first $418$ frames ($80\%$ of the data) for training and the remaining $105$ frames ($20\%$ of the data) for testing. As in the previous example, we apply SVD to the training data and retain only $90\%$ of the total cumulative singular energy, hence reducing the dimensionality to $r = 14$.

For EDMD-kDL, we use the autoencoder formulation. We choose the kernel bandwidth to $\sigma = 5$ and the dictionary size $N = 3 \ts r$. The embedding dimension is set to $8$ for KAE and cKAE after manual hyperparameter tuning. For all methods, we use multi-step prediction for training with $T = 12$.

\begin{figure*}
    \centering
    \includegraphics[width=0.8\linewidth]{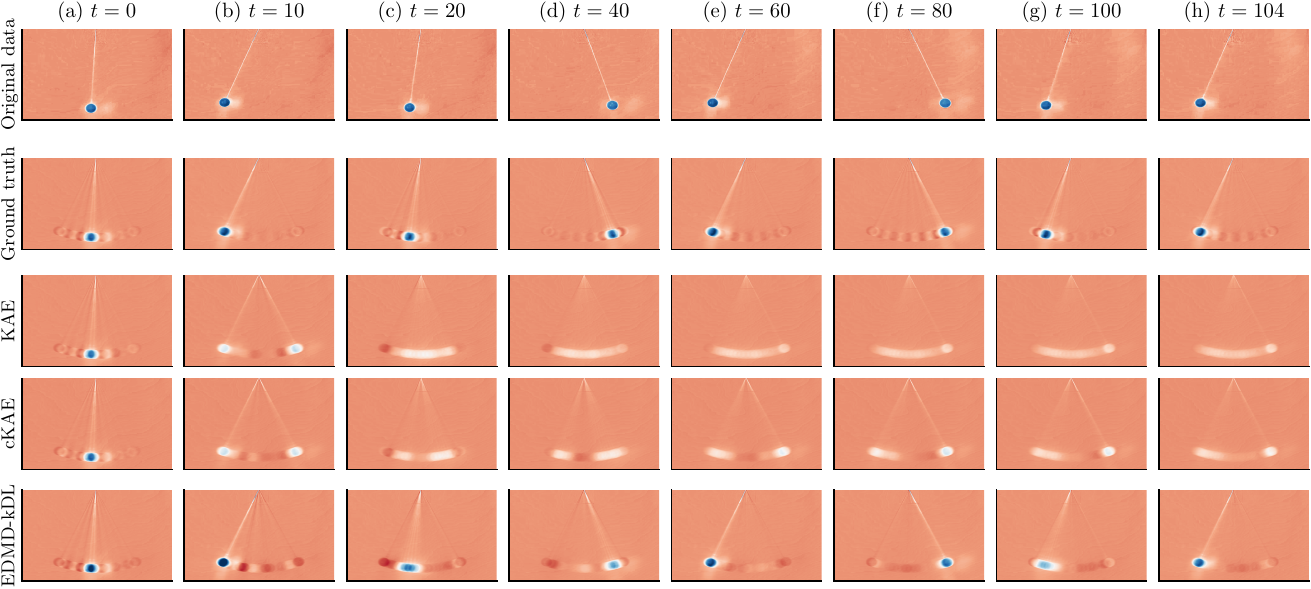}
    \caption{\textbf{Pendulum video data prediction}. Predictions obtained by various Koopman methods in the pixel space. ``Ground truth'' refers to data obtained by projecting the original test data onto the retained SVD modes and reconstructing.}
    \label{fig:pendulum_video_snapshots}
\end{figure*}

Figure~\ref{fig:pendulum_video_snapshots} shows the predictions obtained by each method. We include the reconstructed test data from the SVD projection (the second row) to highlight the error induced by dimensionality reduction, which is independent of the method used to model the temporal evolution in the reduced space. KAE and cKAE drastically fail to properly learn the dynamics of the system. In contrast, EDMD-kDL produces sensible predictions over the entire test horizon. In some cases, the position of the pendulum object is blurred but can still be correctly identified. Moreover, we observe no increase in performance for KAE and cKAE by increasing the embedding dimension.

%%%%%%%%%%%%%%%%%%%%%%%%%%%%%%%%%%%%%%%

\subsection{Global sea-surface temperature forecasting}
As a final example, we consider the task of forecasting sea-surface temperature (SST). The dataset contains $1\ts400$ weekly snapshots of global SST at $1\degree \times 1\degree$ spatial resolution on a $180 \times 360$ grid over the period from $1992$ to $2019$. Of the $64\ts800$ grid points, there are exactly $44\ts219$ valid SST locations. This reanalysis dataset was produced by the National Oceanic and Atmospheric Administration (NOAA); the methodology is described in Ref.~\onlinecite{reynolds2002improved}. We obtain the data from Ref.~\onlinecite{gao2025sparse_dataset} and use the first $1\ts000$ snapshots for training and the remaining $400$ for testing. Since SST values are on different scales across the globe, we first apply min-max scaling at each location so that all the values are contained in $[0, 1]$. Moreover, we remove $720$ rows with constant values over the training period. We then reduce the dimensionality from $43\ts499$ to $r = 19$ by applying SVD and retaining $90\%$ of the cumulative singular energy.

For EDMD-kDL, we set the kernel bandwidth to $\sigma = 10$ and the dictionary size to $N = 2 \ts r$ with a learned decoder. The latent dimension is set to $8$ for KAE and cKAE after manual tuning. We further compare against SINDy-SHRED~\cite{gao2026sparse}, a recently proposed architecture that combines sparse nonlinear system identification~\cite{BPK16} with shallow recurrent decoders~\cite{williams2024sensing}. We use the official SINDy-SHRED implementation available at \href{https://github.com/pyshred-dev/pyshred}{github.com/pyshred-dev/pyshred} with the same settings as in Ref.~\onlinecite{gao2026sparse}, i.e., $250$ sensors with a lag time of $52$ weeks and a total forecasting period of $318$ weeks, corresponding to just over $6$ years.

\begin{figure*}
    \centering
    \includegraphics[width=0.7\linewidth]{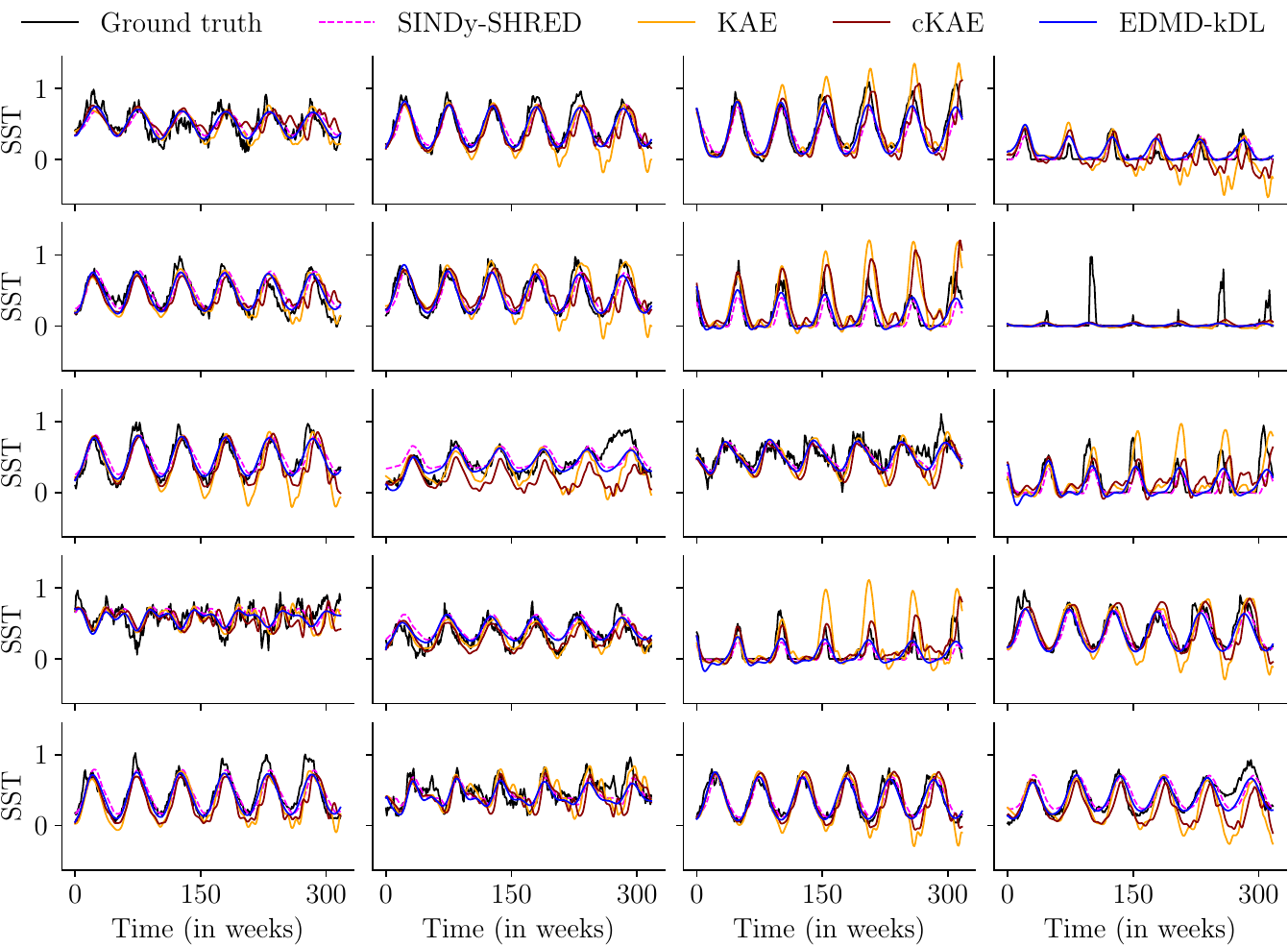}
    \caption{\textbf{Sea surface temperature forecasting}. Scaled predictions at randomly sampled SST locations.}
    \label{fig:sst_prediction_sensors}
\end{figure*}

Figure~\ref{fig:sst_prediction_sensors} shows predictions obtained by different methods for a set of randomly sampled SST locations over the forecast period. EDMD-kDL and SINDy-SHRED produce similar forecasts across the board. The predicted trajectories appear like smoothed versions of the ground truth trajectories. Overall, the forecasts obtained from KAE and cKAE are of lesser quality.

\begin{figure}
    \centering
    \includegraphics[width=0.9\linewidth]{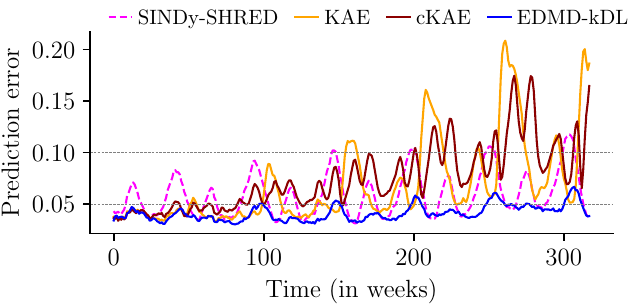}
    \caption{\textbf{Sea surface temperature forecasting}. Relative prediction errors over time.}
    \label{fig:sst_prediction_error}
\end{figure}

\begin{figure*}
    \begin{subfigure}{.48\textwidth}
        \centering
        \includegraphics[width=1.0\linewidth]{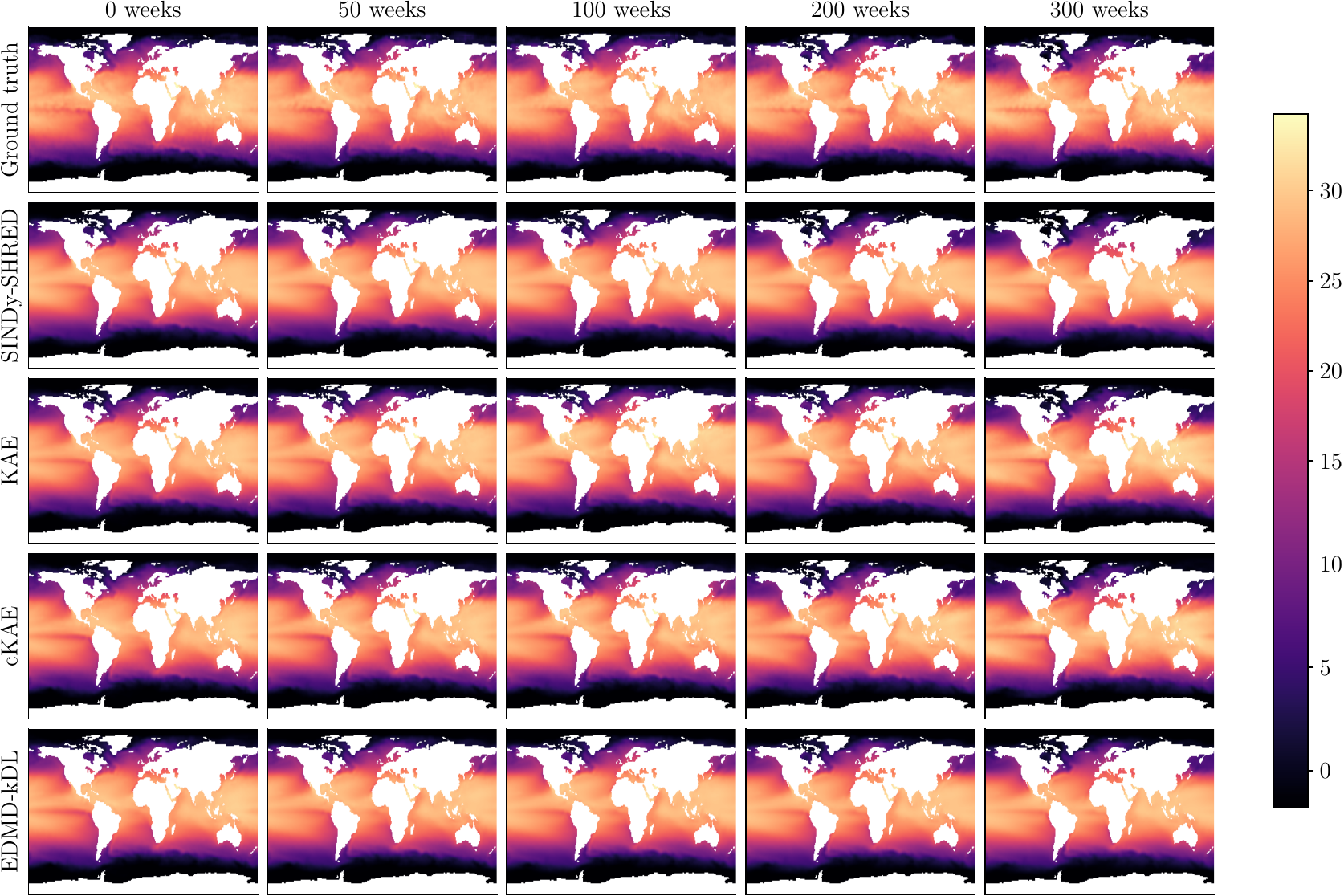}
        \caption{Predictions vs true SST values ($\degree$C).}
        \label{fig:sst_prediction_maps}
    \end{subfigure}
    \begin{subfigure}{.48\textwidth}
        \centering
        \includegraphics[width=1.0\linewidth]{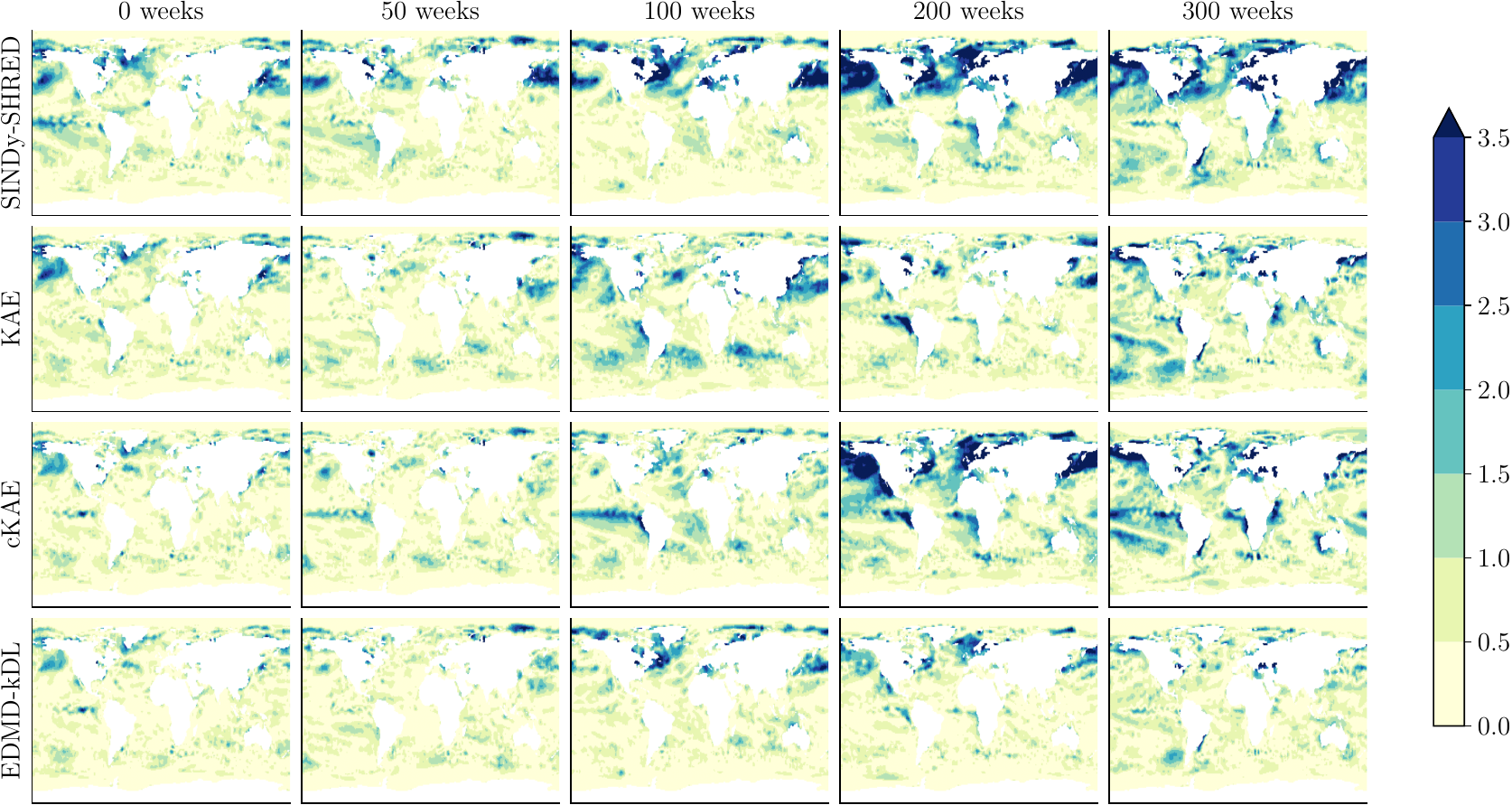}
        \caption{Absolute prediction errors ($\degree$C).}
        \label{fig:sst_prediction_error_maps}
    \end{subfigure}
    \caption{\textbf{Sea surface temperature forecasting}. \textbf{(a)} Predicted global SST maps at specified time steps. \textbf{(b)} Associated absolute errors.}
\end{figure*}

To obtain a more global picture, we unscale the predictions and calculate relative prediction errors based on full snapshots over time. We observe that for the first $100$ weeks, EDMD-kDL, KAE, and cKAE have comparably low error levels, roughly below $5\%$. For this initial phase, SINDy-SHRED has the worst performance overall, see Figure~\ref{fig:sst_prediction_error}. In contrast, for longer term prediction, EDMD-kDL is the only method that maintains a low error level over the entire forecast horizon. The maximum relative error over the $318$-week period is $6.7\%$ for EDMD-kDL, followed by $11.8\%$ for SINDy-SHRED, $17.5\%$ for cKAE, and $20.9\%$ for KAE. Furthermore, we calculate the mean absolute error across SST locations and over time for each method as a summarized performance measure. We obtain $0.51 \degree \text{C}$ for EDMD-kDL, $0.67 \degree \text{C}$ for SINDy-SHRED, $0.83 \degree \text{C}$ for KAE, and $0.88 \degree \text{C}$ for cKAE. This shows that, across the entire prediction horizon, SINDy-SHRED performs better overall compared to KAE and cKAE. Finally, we provide representative predicted spatiotemporal SST maps and the corresponding absolute prediction error maps in Figure~\ref{fig:sst_prediction_maps} and Figure~\ref{fig:sst_prediction_error_maps}, respectively, for illustration purposes.

%%%%%%%%%%%%%%%%%%%%%%%%%%%%%%%%%%%%%%%%
\section{Discussion}
\label{sec:discussion}

In this work, we proposed EDMD-kDL, a novel approach for the data-driven approximation of the Koopman operator that combines the standard EDMD algorithm~\cite{WKR15} with kernel-based autoencoders introduced herein for dictionary learning. EDMD-kDL bridges between standard kernel-based methods for the data-driven spectral decomposition of the Koopman operator~\cite{WRK15, KSM19} and ANN-based autoencoders for learning Koopman embeddings from data~\cite{LDBK17, takeishi2017learning, MPWN18, wehmeyer2018time, yeung2019learning, Otto19, AELM20}. The proposed method is formulated as solving a joint optimization problem over an embedding map in a reproducing kernel Hilbert space and the associated Koopman matrix such that the embedded dynamical system is (approximately) linear. The approach also allows us to learn a linear decoder that reconstructs state space vectors from their embedding space representations. We tackle the resulting function-space optimization problem by introducing pseudo-values for the embedding map at a finite set of collocation points, and reformulating the problem using bilevel optimization. Then, by leveraging the particular structure of RKHSs, notably the representer theorem~\cite{kimeldorf1971some, scholkopf2001generalized}, we obtain a finite-dimensional optimization problem over the pseudo-values, the Koopman matrix, and the decoder matrix, which we solve using standard numerical methods.

We considered four different examples to illustrate the performance of EDMD-kDL in practice: the two-dimensional nonlinear pendulum model with no friction, a simulated fluid flow past a cylinder exhibiting vortex shedding, a video of a pendulum, and global SST. For each experiment, we compare the performance of EDMD-kDL against state-of-the-art ANN-based counterparts on the task of predicting the evolution of the system using learned Koopman models. Across all settings, we find that EDMD-kDL performs comparably or better than the ANN-based approaches. This includes the recently proposed SINDy-SHRED method~\cite{gao2026sparse} for the global SST forecasting example. Notably, the proposed method is able to accurately predict the dynamics of a pendulum directly from video data, a setting in which the ANN-based approaches drastically failed. Similar observations have been made in the context of operator learning~\cite{batlle2024kernel}.

It is worth noting that, although we have focused on deterministic systems in the current work, the proposed approach is directly applicable to stochastic systems. Alternatively, the optimization paradigm can be modified to use the variational formulation, see, e.g., Refs.~\onlinecite{NoNu13, NKPMN14, WuNo20, SKH23}. In this context, it would be of interest to compare EDMD-kDL against VAMPnets~\cite{MPWN18} and time-lagged autoencoders~\cite{wehmeyer2018time}.

Furthermore, we envision multiple potential avenues for future research. Firstly, elucidating the convergence properties and finite-data error bounds for the proposed method -- potentially by combining ideas from the convergence analysis of standard (kernel) EDMD methods~\cite{korda2018convergence, philipp2025error, kohne2025error} and the error analysis of kernel collocation methods for solving nonlinear parametric PDEs~\cite{chen2021solving, batlle2025error} -- would be of great theoretical value. Secondly, our current approach for dealing with high-dimensional systems is to first reduce dimensionality using principal component analysis and then apply EDMD-kDL in the reduced space. Extending the proposed method to work directly with high-dimensional data, for example by jointly learning a nonlinear projection map akin to kernel principal component analysis~\cite{scholkopf1998nonlinear}, is a problem that we would like to investigate in the future. Finally, from a computational standpoint, we observed that starting with a first-order optimizer before switching to L-BFGS when training EDMD-kDL models can substantially reduce the computational cost. We hope to investigate this, and more broadly strategies for efficient training, in the future.

%%%%%%%%%%%%%%%%%%%%%%%%%%%%%%%%%%%%%%%
\begin{acknowledgments}
J.-P.\ was supported by the EPSRC Centre for Doctoral Training in Mathematical Modelling, Analysis and Computation (MAC-MIGS) funded by the UK Engineering and Physical Sciences Research Council (grant EP/S023291/1), Heriot--Watt University and The University of Edinburgh.
\end{acknowledgments}

%%%%%%%%%%%%%%%%%%%%%%%%%%
\section*{Data Availability Statement}
The data that support the findings of this study are available from the corresponding author upon reasonable request.

\section*{References}
\bibliography{edmd_kdl}

\begin{thebibliography}{73}%
\makeatletter
\providecommand \@ifxundefined [1]{%
 \@ifx{#1\undefined}
}%
\providecommand \@ifnum [1]{%
 \ifnum #1\expandafter \@firstoftwo
 \else \expandafter \@secondoftwo
 \fi
}%
\providecommand \@ifx [1]{%
 \ifx #1\expandafter \@firstoftwo
 \else \expandafter \@secondoftwo
 \fi
}%
\providecommand \natexlab [1]{#1}%
\providecommand \enquote  [1]{``#1''}%
\providecommand \bibnamefont  [1]{#1}%
\providecommand \bibfnamefont [1]{#1}%
\providecommand \citenamefont [1]{#1}%
\providecommand \href@noop [0]{\@secondoftwo}%
\providecommand \href [0]{\begingroup \@sanitize@url \@href}%
\providecommand \@href[1]{\@@startlink{#1}\@@href}%
\providecommand \@@href[1]{\endgroup#1\@@endlink}%
\providecommand \@sanitize@url [0]{\catcode `\\12\catcode `\$12\catcode
  `\&12\catcode `\#12\catcode `\^12\catcode `\_12\catcode `\%12\relax}%
\providecommand \@@startlink[1]{}%
\providecommand \@@endlink[0]{}%
\providecommand \url  [0]{\begingroup\@sanitize@url \@url }%
\providecommand \@url [1]{\endgroup\@href {#1}{\urlprefix }}%
\providecommand \urlprefix  [0]{URL }%
\providecommand \Eprint [0]{\href }%
\providecommand \doibase [0]{http://dx.doi.org/}%
\providecommand \selectlanguage [0]{\@gobble}%
\providecommand \bibinfo  [0]{\@secondoftwo}%
\providecommand \bibfield  [0]{\@secondoftwo}%
\providecommand \translation [1]{[#1]}%
\providecommand \BibitemOpen [0]{}%
\providecommand \bibitemStop [0]{}%
\providecommand \bibitemNoStop [0]{.\EOS\space}%
\providecommand \EOS [0]{\spacefactor3000\relax}%
\providecommand \BibitemShut  [1]{\csname bibitem#1\endcsname}%
\let\auto@bib@innerbib\@empty
%</preamble>
\bibitem [{\citenamefont {Koopman}(1931)}]{Ko31}%
  \BibitemOpen
  \bibfield  {author} {\bibinfo {author} {\bibfnamefont {B.~O.}\ \bibnamefont
  {Koopman}},\ }\bibfield  {title} {\enquote {\bibinfo {title} {Hamiltonian
  systems and transformations in {H}ilbert space},}\ }\href {\doibase
  10.1073/pnas.17.5.315} {\bibfield  {journal} {\bibinfo  {journal}
  {Proceedings of the {N}ational {A}cademy of {S}ciences}\ }\textbf {\bibinfo
  {volume} {17}},\ \bibinfo {pages} {315} (\bibinfo {year} {1931})}\BibitemShut
  {NoStop}%
\bibitem [{\citenamefont {Koopman}\ and\ \citenamefont
  {Neumann}(1932)}]{koopman1932dynamical}%
  \BibitemOpen
  \bibfield  {author} {\bibinfo {author} {\bibfnamefont {B.~O.}\ \bibnamefont
  {Koopman}}\ and\ \bibinfo {author} {\bibfnamefont {J.~v.}\ \bibnamefont
  {Neumann}},\ }\bibfield  {title} {\enquote {\bibinfo {title} {Dynamical
  systems of continuous spectra},}\ }\href@noop {} {\bibfield  {journal}
  {\bibinfo  {journal} {Proceedings of the National Academy of Sciences}\
  }\textbf {\bibinfo {volume} {18}},\ \bibinfo {pages} {255--263} (\bibinfo
  {year} {1932})}\BibitemShut {NoStop}%
\bibitem [{\citenamefont {Lasota}\ and\ \citenamefont {Mackey}(1994)}]{LaMa94}%
  \BibitemOpen
  \bibfield  {author} {\bibinfo {author} {\bibfnamefont {A.}~\bibnamefont
  {Lasota}}\ and\ \bibinfo {author} {\bibfnamefont {M.~C.}\ \bibnamefont
  {Mackey}},\ }\href@noop {} {\emph {\bibinfo {title} {Chaos, fractals, and
  noise: Stochastic aspects of dynamics}}},\ \bibinfo {edition} {2nd}\ ed.,\
  \bibinfo {series} {Applied Mathematical Sciences}, Vol.~\bibinfo {volume}
  {97}\ (\bibinfo  {publisher} {Springer},\ \bibinfo {address} {New York},\
  \bibinfo {year} {1994})\BibitemShut {NoStop}%
\bibitem [{\citenamefont {Mezi{\'{c}}}(2005)}]{Mezic05}%
  \BibitemOpen
  \bibfield  {author} {\bibinfo {author} {\bibfnamefont {I.}~\bibnamefont
  {Mezi{\'{c}}}},\ }\bibfield  {title} {\enquote {\bibinfo {title} {Spectral
  properties of dynamical systems, model reduction and decompositions},}\
  }\href {\doibase 10.1007/s11071-005-2824-x} {\bibfield  {journal} {\bibinfo
  {journal} {Nonlinear Dynamics}\ }\textbf {\bibinfo {volume} {41}},\ \bibinfo
  {pages} {309--325} (\bibinfo {year} {2005})}\BibitemShut {NoStop}%
\bibitem [{\citenamefont {Schmid}(2010)}]{Schmid10}%
  \BibitemOpen
  \bibfield  {author} {\bibinfo {author} {\bibfnamefont {P.~J.}\ \bibnamefont
  {Schmid}},\ }\bibfield  {title} {\enquote {\bibinfo {title} {Dynamic mode
  decomposition of numerical and experimental data},}\ }\href {\doibase
  10.1017/S0022112010001217} {\bibfield  {journal} {\bibinfo  {journal}
  {Journal of Fluid Mechanics}\ }\textbf {\bibinfo {volume} {656}},\ \bibinfo
  {pages} {5--28} (\bibinfo {year} {2010})}\BibitemShut {NoStop}%
\bibitem [{\citenamefont {Williams}, \citenamefont {Kevrekidis},\ and\
  \citenamefont {Rowley}(2015)}]{WKR15}%
  \BibitemOpen
  \bibfield  {author} {\bibinfo {author} {\bibfnamefont {M.~O.}\ \bibnamefont
  {Williams}}, \bibinfo {author} {\bibfnamefont {I.~G.}\ \bibnamefont
  {Kevrekidis}}, \ and\ \bibinfo {author} {\bibfnamefont {C.~W.}\ \bibnamefont
  {Rowley}},\ }\bibfield  {title} {\enquote {\bibinfo {title} {A data-driven
  approximation of the {K}oopman operator: Extending dynamic mode
  decomposition},}\ }\href {\doibase 10.1007/s00332-015-9258-5} {\bibfield
  {journal} {\bibinfo  {journal} {Journal of Nonlinear Science}\ }\textbf
  {\bibinfo {volume} {25}},\ \bibinfo {pages} {1307--1346} (\bibinfo {year}
  {2015})}\BibitemShut {NoStop}%
\bibitem [{\citenamefont {Klus}, \citenamefont {Koltai},\ and\ \citenamefont
  {Sch{\"u}tte}(2016)}]{KKS16}%
  \BibitemOpen
  \bibfield  {author} {\bibinfo {author} {\bibfnamefont {S.}~\bibnamefont
  {Klus}}, \bibinfo {author} {\bibfnamefont {P.}~\bibnamefont {Koltai}}, \ and\
  \bibinfo {author} {\bibfnamefont {C.}~\bibnamefont {Sch{\"u}tte}},\
  }\bibfield  {title} {\enquote {\bibinfo {title} {On the numerical
  approximation of the {P}erron--{F}robenius and {K}oopman operator},}\ }\href
  {\doibase 10.3934/jcd.2016003} {\bibfield  {journal} {\bibinfo  {journal}
  {Journal of Computational Dynamics}\ }\textbf {\bibinfo {volume} {3}},\
  \bibinfo {pages} {51--79} (\bibinfo {year} {2016})}\BibitemShut {NoStop}%
\bibitem [{\citenamefont {Li}\ \emph {et~al.}(2017)\citenamefont {Li},
  \citenamefont {Dietrich}, \citenamefont {Bollt},\ and\ \citenamefont
  {Kevrekidis}}]{LDBK17}%
  \BibitemOpen
  \bibfield  {author} {\bibinfo {author} {\bibfnamefont {Q.}~\bibnamefont
  {Li}}, \bibinfo {author} {\bibfnamefont {F.}~\bibnamefont {Dietrich}},
  \bibinfo {author} {\bibfnamefont {E.~M.}\ \bibnamefont {Bollt}}, \ and\
  \bibinfo {author} {\bibfnamefont {I.~G.}\ \bibnamefont {Kevrekidis}},\
  }\bibfield  {title} {\enquote {\bibinfo {title} {Extended dynamic mode
  decomposition with dictionary learning: A data-driven adaptive spectral
  decomposition of the {K}oopman operator},}\ }\href {\doibase
  10.1063/1.4993854} {\bibfield  {journal} {\bibinfo  {journal} {Chaos: An
  Interdisciplinary Journal of Nonlinear Science}\ }\textbf {\bibinfo {volume}
  {27}},\ \bibinfo {pages} {103111} (\bibinfo {year} {2017})}\BibitemShut
  {NoStop}%
\bibitem [{\citenamefont {Klus}\ \emph {et~al.}(2018)\citenamefont {Klus},
  \citenamefont {N\"uske}, \citenamefont {Koltai}, \citenamefont {Wu},
  \citenamefont {Kevrekidis}, \citenamefont {Sch\"utte},\ and\ \citenamefont
  {No\'e}}]{KNKWKSN18}%
  \BibitemOpen
  \bibfield  {author} {\bibinfo {author} {\bibfnamefont {S.}~\bibnamefont
  {Klus}}, \bibinfo {author} {\bibfnamefont {F.}~\bibnamefont {N\"uske}},
  \bibinfo {author} {\bibfnamefont {P.}~\bibnamefont {Koltai}}, \bibinfo
  {author} {\bibfnamefont {H.}~\bibnamefont {Wu}}, \bibinfo {author}
  {\bibfnamefont {I.}~\bibnamefont {Kevrekidis}}, \bibinfo {author}
  {\bibfnamefont {C.}~\bibnamefont {Sch\"utte}}, \ and\ \bibinfo {author}
  {\bibfnamefont {F.}~\bibnamefont {No\'e}},\ }\bibfield  {title} {\enquote
  {\bibinfo {title} {Data-driven model reduction and transfer operator
  approximation},}\ }\href {\doibase 10.1007/s00332-017-9437-7} {\bibfield
  {journal} {\bibinfo  {journal} {Journal of Nonlinear Science}\ }\textbf
  {\bibinfo {volume} {28}},\ \bibinfo {pages} {985--1010} (\bibinfo {year}
  {2018})}\BibitemShut {NoStop}%
\bibitem [{\citenamefont {No{\'e}}\ and\ \citenamefont
  {N{\"u}ske}(2013)}]{NoNu13}%
  \BibitemOpen
  \bibfield  {author} {\bibinfo {author} {\bibfnamefont {F.}~\bibnamefont
  {No{\'e}}}\ and\ \bibinfo {author} {\bibfnamefont {F.}~\bibnamefont
  {N{\"u}ske}},\ }\bibfield  {title} {\enquote {\bibinfo {title} {A variational
  approach to modeling slow processes in stochastic dynamical systems},}\
  }\href@noop {} {\bibfield  {journal} {\bibinfo  {journal} {Multiscale
  Modeling \& Simulation}\ }\textbf {\bibinfo {volume} {11}},\ \bibinfo {pages}
  {635--655} (\bibinfo {year} {2013})}\BibitemShut {NoStop}%
\bibitem [{\citenamefont {N\"uske}\ \emph {et~al.}(2014)\citenamefont
  {N\"uske}, \citenamefont {Keller}, \citenamefont {P\'erez-Hern\'andez},
  \citenamefont {Mey},\ and\ \citenamefont {No\'e}}]{NKPMN14}%
  \BibitemOpen
  \bibfield  {author} {\bibinfo {author} {\bibfnamefont {F.}~\bibnamefont
  {N\"uske}}, \bibinfo {author} {\bibfnamefont {B.~G.}\ \bibnamefont {Keller}},
  \bibinfo {author} {\bibfnamefont {G.}~\bibnamefont {P\'erez-Hern\'andez}},
  \bibinfo {author} {\bibfnamefont {A.~S. J.~S.}\ \bibnamefont {Mey}}, \ and\
  \bibinfo {author} {\bibfnamefont {F.}~\bibnamefont {No\'e}},\ }\bibfield
  {title} {\enquote {\bibinfo {title} {Variational approach to molecular
  kinetics},}\ }\href@noop {} {\bibfield  {journal} {\bibinfo  {journal}
  {Journal of Chemical Theory and Computation}\ }\textbf {\bibinfo {volume}
  {10}},\ \bibinfo {pages} {1739--1752} (\bibinfo {year} {2014})}\BibitemShut
  {NoStop}%
\bibitem [{\citenamefont {Molgedey}\ and\ \citenamefont
  {Schuster}(1994)}]{MS94}%
  \BibitemOpen
  \bibfield  {author} {\bibinfo {author} {\bibfnamefont {L.}~\bibnamefont
  {Molgedey}}\ and\ \bibinfo {author} {\bibfnamefont {H.~G.}\ \bibnamefont
  {Schuster}},\ }\bibfield  {title} {\enquote {\bibinfo {title} {Separation of
  a mixture of independent signals using time delayed correlations},}\
  }\href@noop {} {\bibfield  {journal} {\bibinfo  {journal} {Physical Review
  Letters}\ }\textbf {\bibinfo {volume} {72}},\ \bibinfo {pages} {3634--3637}
  (\bibinfo {year} {1994})}\BibitemShut {NoStop}%
\bibitem [{\citenamefont {Mezi{\'c}}(2013)}]{mezic2013analysis}%
  \BibitemOpen
  \bibfield  {author} {\bibinfo {author} {\bibfnamefont {I.}~\bibnamefont
  {Mezi{\'c}}},\ }\bibfield  {title} {\enquote {\bibinfo {title} {Analysis of
  fluid flows via spectral properties of the {K}oopman operator},}\ }\href@noop
  {} {\bibfield  {journal} {\bibinfo  {journal} {Annual Review of Fluid
  Mechanics}\ }\textbf {\bibinfo {volume} {45}},\ \bibinfo {pages} {357--378}
  (\bibinfo {year} {2013})}\BibitemShut {NoStop}%
\bibitem [{\citenamefont {Tu}\ \emph {et~al.}(2014)\citenamefont {Tu},
  \citenamefont {Rowley}, \citenamefont {Luchtenburg}, \citenamefont
  {Brunton},\ and\ \citenamefont {Kutz}}]{TRLBK14}%
  \BibitemOpen
  \bibfield  {author} {\bibinfo {author} {\bibfnamefont {J.~H.}\ \bibnamefont
  {Tu}}, \bibinfo {author} {\bibfnamefont {C.~W.}\ \bibnamefont {Rowley}},
  \bibinfo {author} {\bibfnamefont {D.~M.}\ \bibnamefont {Luchtenburg}},
  \bibinfo {author} {\bibfnamefont {S.~L.}\ \bibnamefont {Brunton}}, \ and\
  \bibinfo {author} {\bibfnamefont {J.~N.}\ \bibnamefont {Kutz}},\ }\bibfield
  {title} {\enquote {\bibinfo {title} {On dynamic mode decomposition: {T}heory
  and applications},}\ }\href@noop {} {\bibfield  {journal} {\bibinfo
  {journal} {Journal of Computational Dynamics}\ }\textbf {\bibinfo {volume}
  {1}} (\bibinfo {year} {2014})}\BibitemShut {NoStop}%
\bibitem [{\citenamefont {Otto}\ and\ \citenamefont
  {Rowley}(2021)}]{otto2021koopman}%
  \BibitemOpen
  \bibfield  {author} {\bibinfo {author} {\bibfnamefont {S.~E.}\ \bibnamefont
  {Otto}}\ and\ \bibinfo {author} {\bibfnamefont {C.~W.}\ \bibnamefont
  {Rowley}},\ }\bibfield  {title} {\enquote {\bibinfo {title} {Koopman
  operators for estimation and control of dynamical systems},}\ }\href@noop {}
  {\bibfield  {journal} {\bibinfo  {journal} {Annual Review of Control,
  Robotics, and Autonomous Systems}\ }\textbf {\bibinfo {volume} {4}},\
  \bibinfo {pages} {59--87} (\bibinfo {year} {2021})}\BibitemShut {NoStop}%
\bibitem [{\citenamefont {Schmid}(2022)}]{schmid2022dynamic}%
  \BibitemOpen
  \bibfield  {author} {\bibinfo {author} {\bibfnamefont {P.~J.}\ \bibnamefont
  {Schmid}},\ }\bibfield  {title} {\enquote {\bibinfo {title} {Dynamic mode
  decomposition and its variants},}\ }\href@noop {} {\bibfield  {journal}
  {\bibinfo  {journal} {Annual Review of Fluid Mechanics}\ }\textbf {\bibinfo
  {volume} {54}},\ \bibinfo {pages} {225--254} (\bibinfo {year}
  {2022})}\BibitemShut {NoStop}%
\bibitem [{\citenamefont {Colbrook}(2024)}]{colbrook2024multiverse}%
  \BibitemOpen
  \bibfield  {author} {\bibinfo {author} {\bibfnamefont {M.~J.}\ \bibnamefont
  {Colbrook}},\ }\bibfield  {title} {\enquote {\bibinfo {title} {The multiverse
  of dynamic mode decomposition algorithms},}\ }in\ \href@noop {} {\emph
  {\bibinfo {booktitle} {Handbook of numerical analysis}}},\ Vol.~\bibinfo
  {volume} {25}\ (\bibinfo  {publisher} {Elsevier},\ \bibinfo {year} {2024})\
  pp.\ \bibinfo {pages} {127--230}\BibitemShut {NoStop}%
\bibitem [{\citenamefont {Mauroy}\ and\ \citenamefont
  {Goncalves}(2016)}]{MauGon16}%
  \BibitemOpen
  \bibfield  {author} {\bibinfo {author} {\bibfnamefont {A.}~\bibnamefont
  {Mauroy}}\ and\ \bibinfo {author} {\bibfnamefont {J.}~\bibnamefont
  {Goncalves}},\ }\bibfield  {title} {\enquote {\bibinfo {title} {Linear
  identification of nonlinear systems: {A} lifting technique based on the
  {K}oopman operator},}\ }in\ \href {\doibase 10.1109/CDC.2016.7799269} {\emph
  {\bibinfo {booktitle} {2016 IEEE 55th Conference on Decision and Control
  (CDC)}}}\ (\bibinfo {year} {2016})\ pp.\ \bibinfo {pages}
  {6500--6505}\BibitemShut {NoStop}%
\bibitem [{\citenamefont {Bruder}, \citenamefont {Remy},\ and\ \citenamefont
  {Vasudevan}(2019)}]{bruder2019nonlinear}%
  \BibitemOpen
  \bibfield  {author} {\bibinfo {author} {\bibfnamefont {D.}~\bibnamefont
  {Bruder}}, \bibinfo {author} {\bibfnamefont {C.~D.}\ \bibnamefont {Remy}}, \
  and\ \bibinfo {author} {\bibfnamefont {R.}~\bibnamefont {Vasudevan}},\
  }\bibfield  {title} {\enquote {\bibinfo {title} {Nonlinear system
  identification of soft robot dynamics using {K}oopman operator theory},}\
  }in\ \href@noop {} {\emph {\bibinfo {booktitle} {2019 International
  Conference on Robotics and Automation (ICRA)}}}\ (\bibinfo {organization}
  {IEEE},\ \bibinfo {year} {2019})\ pp.\ \bibinfo {pages}
  {6244--6250}\BibitemShut {NoStop}%
\bibitem [{\citenamefont {Klus}\ \emph {et~al.}(2020)\citenamefont {Klus},
  \citenamefont {N\"uske}, \citenamefont {Peitz}, \citenamefont {Niemann},
  \citenamefont {Clementi},\ and\ \citenamefont {Sch\"utte}}]{KNPNCS20}%
  \BibitemOpen
  \bibfield  {author} {\bibinfo {author} {\bibfnamefont {S.}~\bibnamefont
  {Klus}}, \bibinfo {author} {\bibfnamefont {F.}~\bibnamefont {N\"uske}},
  \bibinfo {author} {\bibfnamefont {S.}~\bibnamefont {Peitz}}, \bibinfo
  {author} {\bibfnamefont {J.-H.}\ \bibnamefont {Niemann}}, \bibinfo {author}
  {\bibfnamefont {C.}~\bibnamefont {Clementi}}, \ and\ \bibinfo {author}
  {\bibfnamefont {C.}~\bibnamefont {Sch\"utte}},\ }\bibfield  {title} {\enquote
  {\bibinfo {title} {Data-driven approximation of the {K}oopman generator:
  {M}odel reduction, system identification, and control},}\ }\href {\doibase
  10.1016/j.physd.2020.132416} {\bibfield  {journal} {\bibinfo  {journal}
  {Physica D: Nonlinear Phenomena}\ }\textbf {\bibinfo {volume} {406}},\
  \bibinfo {pages} {132416} (\bibinfo {year} {2020})}\BibitemShut {NoStop}%
\bibitem [{\citenamefont {Peitz}\ and\ \citenamefont {Klus}(2019)}]{PeiKlul19}%
  \BibitemOpen
  \bibfield  {author} {\bibinfo {author} {\bibfnamefont {S.}~\bibnamefont
  {Peitz}}\ and\ \bibinfo {author} {\bibfnamefont {S.}~\bibnamefont {Klus}},\
  }\bibfield  {title} {\enquote {\bibinfo {title} {{K}oopman operator-based
  model reduction for switched-system control of {PDE}s},}\ }\href {\doibase
  10.1016/j.automatica.2019.05.016} {\bibfield  {journal} {\bibinfo  {journal}
  {Automatica}\ }\textbf {\bibinfo {volume} {106}},\ \bibinfo {pages}
  {184--191} (\bibinfo {year} {2019})}\BibitemShut {NoStop}%
\bibitem [{\citenamefont {Nateghi}\ and\ \citenamefont
  {N\"uske}(2025)}]{nateghi2025kinetically}%
  \BibitemOpen
  \bibfield  {author} {\bibinfo {author} {\bibfnamefont {V.}~\bibnamefont
  {Nateghi}}\ and\ \bibinfo {author} {\bibfnamefont {F.}~\bibnamefont
  {N\"uske}},\ }\bibfield  {title} {\enquote {\bibinfo {title} {Kinetically
  consistent coarse graining using kernel-based extended dynamic mode
  decomposition},}\ }\href@noop {} {\bibfield  {journal} {\bibinfo  {journal}
  {Journal of Chemical Theory and Computation}\ }\textbf {\bibinfo {volume}
  {21}},\ \bibinfo {pages} {7236--7248} (\bibinfo {year} {2025})}\BibitemShut
  {NoStop}%
\bibitem [{\citenamefont {Korda}\ and\ \citenamefont
  {Mezi{\'{c}}}(2018)}]{KorMez16}%
  \BibitemOpen
  \bibfield  {author} {\bibinfo {author} {\bibfnamefont {M.}~\bibnamefont
  {Korda}}\ and\ \bibinfo {author} {\bibfnamefont {I.}~\bibnamefont
  {Mezi{\'{c}}}},\ }\bibfield  {title} {\enquote {\bibinfo {title} {Linear
  predictors for nonlinear dynamical systems: {K}oopman operator meets model
  predictive control},}\ }\href {\doibase 10.1016/j.automatica.2018.03.046}
  {\bibfield  {journal} {\bibinfo  {journal} {Automatica}\ }\textbf {\bibinfo
  {volume} {93}},\ \bibinfo {pages} {149--160} (\bibinfo {year}
  {2018})}\BibitemShut {NoStop}%
\bibitem [{\citenamefont {Surana}(2016)}]{surana2016koopman}%
  \BibitemOpen
  \bibfield  {author} {\bibinfo {author} {\bibfnamefont {A.}~\bibnamefont
  {Surana}},\ }\bibfield  {title} {\enquote {\bibinfo {title} {Koopman operator
  based observer synthesis for control-affine nonlinear systems},}\ }in\
  \href@noop {} {\emph {\bibinfo {booktitle} {2016 IEEE 55th Conference on
  Decision and Control (CDC)}}}\ (\bibinfo {organization} {IEEE},\ \bibinfo
  {year} {2016})\ pp.\ \bibinfo {pages} {6492--6499}\BibitemShut {NoStop}%
\bibitem [{\citenamefont {Goswami}\ and\ \citenamefont
  {Paley}(2017)}]{goswami2017global}%
  \BibitemOpen
  \bibfield  {author} {\bibinfo {author} {\bibfnamefont {D.}~\bibnamefont
  {Goswami}}\ and\ \bibinfo {author} {\bibfnamefont {D.~A.}\ \bibnamefont
  {Paley}},\ }\bibfield  {title} {\enquote {\bibinfo {title} {Global
  bilinearization and controllability of control-affine nonlinear systems: A
  {K}oopman spectral approach},}\ }in\ \href@noop {} {\emph {\bibinfo
  {booktitle} {2017 IEEE 56th Annual Conference on Decision and Control
  (CDC)}}}\ (\bibinfo {organization} {IEEE},\ \bibinfo {year} {2017})\ pp.\
  \bibinfo {pages} {6107--6112}\BibitemShut {NoStop}%
\bibitem [{\citenamefont {Arbabi}, \citenamefont {Korda},\ and\ \citenamefont
  {Mezi{\'c}}(2018)}]{arbabi2018data}%
  \BibitemOpen
  \bibfield  {author} {\bibinfo {author} {\bibfnamefont {H.}~\bibnamefont
  {Arbabi}}, \bibinfo {author} {\bibfnamefont {M.}~\bibnamefont {Korda}}, \
  and\ \bibinfo {author} {\bibfnamefont {I.}~\bibnamefont {Mezi{\'c}}},\
  }\bibfield  {title} {\enquote {\bibinfo {title} {A data-driven {K}oopman
  model predictive control framework for nonlinear partial differential
  equations},}\ }in\ \href@noop {} {\emph {\bibinfo {booktitle} {2018 IEEE
  Conference on Decision and Control (CDC)}}}\ (\bibinfo {organization}
  {IEEE},\ \bibinfo {year} {2018})\ pp.\ \bibinfo {pages}
  {6409--6414}\BibitemShut {NoStop}%
\bibitem [{\citenamefont {Bevanda}, \citenamefont {Sosnowski},\ and\
  \citenamefont {Hirche}(2021)}]{BSH21}%
  \BibitemOpen
  \bibfield  {author} {\bibinfo {author} {\bibfnamefont {P.}~\bibnamefont
  {Bevanda}}, \bibinfo {author} {\bibfnamefont {S.}~\bibnamefont {Sosnowski}},
  \ and\ \bibinfo {author} {\bibfnamefont {S.}~\bibnamefont {Hirche}},\
  }\bibfield  {title} {\enquote {\bibinfo {title} {{K}oopman operator dynamical
  models: Learning, analysis and control},}\ }\href {\doibase
  doi.org/10.1016/j.arcontrol.2021.09.002} {\bibfield  {journal} {\bibinfo
  {journal} {Annual Reviews in Control}\ }\textbf {\bibinfo {volume} {52}},\
  \bibinfo {pages} {197--212} (\bibinfo {year} {2021})}\BibitemShut {NoStop}%
\bibitem [{\citenamefont {Hua}\ \emph {et~al.}(2017)\citenamefont {Hua},
  \citenamefont {Noorian}, \citenamefont {Moss}, \citenamefont {Leong},\ and\
  \citenamefont {Gunaratne}}]{HNDLG17}%
  \BibitemOpen
  \bibfield  {author} {\bibinfo {author} {\bibfnamefont {J.}~\bibnamefont
  {Hua}}, \bibinfo {author} {\bibfnamefont {F.}~\bibnamefont {Noorian}},
  \bibinfo {author} {\bibfnamefont {D.}~\bibnamefont {Moss}}, \bibinfo {author}
  {\bibfnamefont {P.~H.~W.}\ \bibnamefont {Leong}}, \ and\ \bibinfo {author}
  {\bibfnamefont {G.~H.}\ \bibnamefont {Gunaratne}},\ }\bibfield  {title}
  {\enquote {\bibinfo {title} {High-dimensional time series prediction using
  kernel-based {K}oopman mode regression},}\ }\href {\doibase
  10.1007/s11071-017-3764-y} {\bibfield  {journal} {\bibinfo  {journal}
  {Nonlinear Dynamics}\ }\textbf {\bibinfo {volume} {90}},\ \bibinfo {pages}
  {1785--1806} (\bibinfo {year} {2017})}\BibitemShut {NoStop}%
\bibitem [{\citenamefont {Giannakis}(2019)}]{Gia19}%
  \BibitemOpen
  \bibfield  {author} {\bibinfo {author} {\bibfnamefont {D.}~\bibnamefont
  {Giannakis}},\ }\bibfield  {title} {\enquote {\bibinfo {title} {Data-driven
  spectral decomposition and forecasting of ergodic dynamical systems},}\
  }\href {\doibase 10.1016/j.acha.2017.09.001} {\bibfield  {journal} {\bibinfo
  {journal} {Applied and Computational Harmonic Analysis}\ }\textbf {\bibinfo
  {volume} {47}},\ \bibinfo {pages} {338--396} (\bibinfo {year}
  {2019})}\BibitemShut {NoStop}%
\bibitem [{\citenamefont {Azencot}\ \emph {et~al.}(2020)\citenamefont
  {Azencot}, \citenamefont {Erichson}, \citenamefont {Lin},\ and\ \citenamefont
  {Mahoney}}]{AELM20}%
  \BibitemOpen
  \bibfield  {author} {\bibinfo {author} {\bibfnamefont {O.}~\bibnamefont
  {Azencot}}, \bibinfo {author} {\bibfnamefont {N.~B.}\ \bibnamefont
  {Erichson}}, \bibinfo {author} {\bibfnamefont {V.}~\bibnamefont {Lin}}, \
  and\ \bibinfo {author} {\bibfnamefont {M.}~\bibnamefont {Mahoney}},\
  }\bibfield  {title} {\enquote {\bibinfo {title} {Forecasting sequential data
  using consistent {K}oopman autoencoders},}\ }in\ \href@noop {} {\emph
  {\bibinfo {booktitle} {International Conference on Machine Learning}}}\
  (\bibinfo {organization} {PMLR},\ \bibinfo {year} {2020})\ pp.\ \bibinfo
  {pages} {475--485}\BibitemShut {NoStop}%
\bibitem [{\citenamefont {Lange}, \citenamefont {Brunton},\ and\ \citenamefont
  {Kutz}(2021)}]{LBK21}%
  \BibitemOpen
  \bibfield  {author} {\bibinfo {author} {\bibfnamefont {H.}~\bibnamefont
  {Lange}}, \bibinfo {author} {\bibfnamefont {S.~L.}\ \bibnamefont {Brunton}},
  \ and\ \bibinfo {author} {\bibfnamefont {J.~N.}\ \bibnamefont {Kutz}},\
  }\bibfield  {title} {\enquote {\bibinfo {title} {From {F}ourier to {K}oopman:
  Spectral methods for long-term time series prediction},}\ }\href@noop {}
  {\bibfield  {journal} {\bibinfo  {journal} {Journal of Machine Learning
  Research}\ }\textbf {\bibinfo {volume} {22}},\ \bibinfo {pages} {1--38}
  (\bibinfo {year} {2021})}\BibitemShut {NoStop}%
\bibitem [{\citenamefont {Bevanda}\ \emph {et~al.}(2023)\citenamefont
  {Bevanda}, \citenamefont {Beier}, \citenamefont {Lederer}, \citenamefont
  {Sosnowski}, \citenamefont {H{\"u}llermeier},\ and\ \citenamefont
  {Hirche}}]{bevanda2023koopman}%
  \BibitemOpen
  \bibfield  {author} {\bibinfo {author} {\bibfnamefont {P.}~\bibnamefont
  {Bevanda}}, \bibinfo {author} {\bibfnamefont {M.}~\bibnamefont {Beier}},
  \bibinfo {author} {\bibfnamefont {A.}~\bibnamefont {Lederer}}, \bibinfo
  {author} {\bibfnamefont {S.}~\bibnamefont {Sosnowski}}, \bibinfo {author}
  {\bibfnamefont {E.}~\bibnamefont {H{\"u}llermeier}}, \ and\ \bibinfo {author}
  {\bibfnamefont {S.}~\bibnamefont {Hirche}},\ }\bibfield  {title} {\enquote
  {\bibinfo {title} {Koopman kernel regression},}\ }\href@noop {} {\bibfield
  {journal} {\bibinfo  {journal} {Advances in Neural Information Processing
  Systems}\ }\textbf {\bibinfo {volume} {36}},\ \bibinfo {pages} {16207--16221}
  (\bibinfo {year} {2023})}\BibitemShut {NoStop}%
\bibitem [{\citenamefont {Takeishi}, \citenamefont {Kawahara},\ and\
  \citenamefont {Yairi}(2017)}]{takeishi2017learning}%
  \BibitemOpen
  \bibfield  {author} {\bibinfo {author} {\bibfnamefont {N.}~\bibnamefont
  {Takeishi}}, \bibinfo {author} {\bibfnamefont {Y.}~\bibnamefont {Kawahara}},
  \ and\ \bibinfo {author} {\bibfnamefont {T.}~\bibnamefont {Yairi}},\
  }\bibfield  {title} {\enquote {\bibinfo {title} {Learning {K}oopman invariant
  subspaces for dynamic mode decomposition},}\ }\href@noop {} {\bibfield
  {journal} {\bibinfo  {journal} {Advances in neural information processing
  systems}\ }\textbf {\bibinfo {volume} {30}} (\bibinfo {year}
  {2017})}\BibitemShut {NoStop}%
\bibitem [{\citenamefont {Mardt}\ \emph {et~al.}(2018)\citenamefont {Mardt},
  \citenamefont {Pasquali}, \citenamefont {Wu},\ and\ \citenamefont
  {No{\'e}}}]{MPWN18}%
  \BibitemOpen
  \bibfield  {author} {\bibinfo {author} {\bibfnamefont {A.}~\bibnamefont
  {Mardt}}, \bibinfo {author} {\bibfnamefont {L.}~\bibnamefont {Pasquali}},
  \bibinfo {author} {\bibfnamefont {H.}~\bibnamefont {Wu}}, \ and\ \bibinfo
  {author} {\bibfnamefont {F.}~\bibnamefont {No{\'e}}},\ }\bibfield  {title}
  {\enquote {\bibinfo {title} {{VAMPnets} for deep learning of molecular
  kinetics},}\ }\href {\doibase 10.1038/s41467-017-02388-1} {\bibfield
  {journal} {\bibinfo  {journal} {Nature Communications}\ }\textbf {\bibinfo
  {volume} {9}} (\bibinfo {year} {2018}),\
  10.1038/s41467-017-02388-1}\BibitemShut {NoStop}%
\bibitem [{\citenamefont {Wehmeyer}\ and\ \citenamefont
  {No{\'e}}(2018)}]{wehmeyer2018time}%
  \BibitemOpen
  \bibfield  {author} {\bibinfo {author} {\bibfnamefont {C.}~\bibnamefont
  {Wehmeyer}}\ and\ \bibinfo {author} {\bibfnamefont {F.}~\bibnamefont
  {No{\'e}}},\ }\bibfield  {title} {\enquote {\bibinfo {title} {Time-lagged
  autoencoders: Deep learning of slow collective variables for molecular
  kinetics},}\ }\href@noop {} {\bibfield  {journal} {\bibinfo  {journal} {The
  Journal of chemical physics}\ }\textbf {\bibinfo {volume} {148}} (\bibinfo
  {year} {2018})}\BibitemShut {NoStop}%
\bibitem [{\citenamefont {Yeung}, \citenamefont {Kundu},\ and\ \citenamefont
  {Hodas}(2019)}]{yeung2019learning}%
  \BibitemOpen
  \bibfield  {author} {\bibinfo {author} {\bibfnamefont {E.}~\bibnamefont
  {Yeung}}, \bibinfo {author} {\bibfnamefont {S.}~\bibnamefont {Kundu}}, \ and\
  \bibinfo {author} {\bibfnamefont {N.}~\bibnamefont {Hodas}},\ }\bibfield
  {title} {\enquote {\bibinfo {title} {Learning deep neural network
  representations for {K}oopman operators of nonlinear dynamical systems},}\
  }in\ \href@noop {} {\emph {\bibinfo {booktitle} {2019 American Control
  Conference (ACC)}}}\ (\bibinfo {organization} {IEEE},\ \bibinfo {year}
  {2019})\ pp.\ \bibinfo {pages} {4832--4839}\BibitemShut {NoStop}%
\bibitem [{\citenamefont {Otto}\ and\ \citenamefont {Rowley}(2019)}]{Otto19}%
  \BibitemOpen
  \bibfield  {author} {\bibinfo {author} {\bibfnamefont {S.~E.}\ \bibnamefont
  {Otto}}\ and\ \bibinfo {author} {\bibfnamefont {C.~W.}\ \bibnamefont
  {Rowley}},\ }\bibfield  {title} {\enquote {\bibinfo {title}
  {Linearly-recurrent autoencoder networks for learning dynamics},}\ }\href
  {\doibase 10.1137/18M1177846} {\bibfield  {journal} {\bibinfo  {journal}
  {SIAM Journal on Applied Dynamical Systems}\ }\textbf {\bibinfo {volume}
  {18}},\ \bibinfo {pages} {558--593} (\bibinfo {year} {2019})}\BibitemShut
  {NoStop}%
\bibitem [{\citenamefont {Tabish}, \citenamefont {Leimkuhler},\ and\
  \citenamefont {Klus}(2025)}]{TLK25}%
  \BibitemOpen
  \bibfield  {author} {\bibinfo {author} {\bibfnamefont {M.}~\bibnamefont
  {Tabish}}, \bibinfo {author} {\bibfnamefont {B.}~\bibnamefont {Leimkuhler}},
  \ and\ \bibinfo {author} {\bibfnamefont {S.}~\bibnamefont {Klus}},\
  }\href@noop {} {\enquote {\bibinfo {title} {How deep is your network? {D}eep
  vs.\ shallow learning of transfer operators},}\ } (\bibinfo {year} {2025}),\
  \Eprint {http://arxiv.org/abs/2509.19930} {arXiv:2509.19930} \BibitemShut
  {NoStop}%
\bibitem [{\citenamefont {Erichson}, \citenamefont {Muehlebach},\ and\
  \citenamefont {Mahoney}(2019)}]{erichson2019physics}%
  \BibitemOpen
  \bibfield  {author} {\bibinfo {author} {\bibfnamefont {N.~B.}\ \bibnamefont
  {Erichson}}, \bibinfo {author} {\bibfnamefont {M.}~\bibnamefont
  {Muehlebach}}, \ and\ \bibinfo {author} {\bibfnamefont {M.~W.}\ \bibnamefont
  {Mahoney}},\ }\bibfield  {title} {\enquote {\bibinfo {title}
  {Physics-informed autoencoders for {L}yapunov-stable fluid flow
  prediction},}\ }\href@noop {} {\bibfield  {journal} {\bibinfo  {journal}
  {arXiv preprint arXiv:1905.10866}\ } (\bibinfo {year} {2019})}\BibitemShut
  {NoStop}%
\bibitem [{\citenamefont {Williams}, \citenamefont {Rowley},\ and\
  \citenamefont {Kevrekidis}(2015)}]{WRK15}%
  \BibitemOpen
  \bibfield  {author} {\bibinfo {author} {\bibfnamefont {M.~O.}\ \bibnamefont
  {Williams}}, \bibinfo {author} {\bibfnamefont {C.~W.}\ \bibnamefont
  {Rowley}}, \ and\ \bibinfo {author} {\bibfnamefont {I.~G.}\ \bibnamefont
  {Kevrekidis}},\ }\bibfield  {title} {\enquote {\bibinfo {title} {A
  kernel-based method for data-driven {K}oopman spectral analysis},}\ }\href
  {\doibase 10.3934/jcd.2015005} {\bibfield  {journal} {\bibinfo  {journal}
  {Journal of Computational Dynamics}\ }\textbf {\bibinfo {volume} {2}},\
  \bibinfo {pages} {247--265} (\bibinfo {year} {2015})}\BibitemShut {NoStop}%
\bibitem [{\citenamefont {Klus}, \citenamefont {Schuster},\ and\ \citenamefont
  {Muandet}(2019)}]{KSM19}%
  \BibitemOpen
  \bibfield  {author} {\bibinfo {author} {\bibfnamefont {S.}~\bibnamefont
  {Klus}}, \bibinfo {author} {\bibfnamefont {I.}~\bibnamefont {Schuster}}, \
  and\ \bibinfo {author} {\bibfnamefont {K.}~\bibnamefont {Muandet}},\
  }\bibfield  {title} {\enquote {\bibinfo {title} {Eigendecompositions of
  transfer operators in reproducing kernel {H}ilbert spaces},}\ }\href
  {\doibase 10.1007/s00332-019-09574-z} {\bibfield  {journal} {\bibinfo
  {journal} {Journal of Nonlinear Science}\ } (\bibinfo {year} {2019}),\
  10.1007/s00332-019-09574-z}\BibitemShut {NoStop}%
\bibitem [{\citenamefont {Cybenko}(1989)}]{cybenko1989approximation}%
  \BibitemOpen
  \bibfield  {author} {\bibinfo {author} {\bibfnamefont {G.}~\bibnamefont
  {Cybenko}},\ }\bibfield  {title} {\enquote {\bibinfo {title} {Approximation
  by superpositions of a sigmoidal function},}\ }\href@noop {} {\bibfield
  {journal} {\bibinfo  {journal} {Mathematics of control, signals and systems}\
  }\textbf {\bibinfo {volume} {2}},\ \bibinfo {pages} {303--314} (\bibinfo
  {year} {1989})}\BibitemShut {NoStop}%
\bibitem [{\citenamefont {Micchelli}, \citenamefont {Xu},\ and\ \citenamefont
  {Zhang}(2006)}]{micchelli2006universal}%
  \BibitemOpen
  \bibfield  {author} {\bibinfo {author} {\bibfnamefont {C.~A.}\ \bibnamefont
  {Micchelli}}, \bibinfo {author} {\bibfnamefont {Y.}~\bibnamefont {Xu}}, \
  and\ \bibinfo {author} {\bibfnamefont {H.}~\bibnamefont {Zhang}},\ }\bibfield
   {title} {\enquote {\bibinfo {title} {Universal kernels.}}\ }\href@noop {}
  {\bibfield  {journal} {\bibinfo  {journal} {Journal of Machine Learning
  Research}\ }\textbf {\bibinfo {volume} {7}} (\bibinfo {year}
  {2006})}\BibitemShut {NoStop}%
\bibitem [{\citenamefont {Sriperumbudur}, \citenamefont {Fukumizu},\ and\
  \citenamefont {Lanckriet}(2011)}]{sriperumbudur2011universality}%
  \BibitemOpen
  \bibfield  {author} {\bibinfo {author} {\bibfnamefont {B.~K.}\ \bibnamefont
  {Sriperumbudur}}, \bibinfo {author} {\bibfnamefont {K.}~\bibnamefont
  {Fukumizu}}, \ and\ \bibinfo {author} {\bibfnamefont {G.~R.}\ \bibnamefont
  {Lanckriet}},\ }\bibfield  {title} {\enquote {\bibinfo {title} {Universality,
  characteristic kernels and {RKHS} embedding of measures.}}\ }\href@noop {}
  {\bibfield  {journal} {\bibinfo  {journal} {Journal of Machine Learning
  Research}\ }\textbf {\bibinfo {volume} {12}} (\bibinfo {year}
  {2011})}\BibitemShut {NoStop}%
\bibitem [{\citenamefont {Schaback}\ and\ \citenamefont
  {Wendland}(2006)}]{schaback2006kernel}%
  \BibitemOpen
  \bibfield  {author} {\bibinfo {author} {\bibfnamefont {R.}~\bibnamefont
  {Schaback}}\ and\ \bibinfo {author} {\bibfnamefont {H.}~\bibnamefont
  {Wendland}},\ }\bibfield  {title} {\enquote {\bibinfo {title} {Kernel
  techniques: from machine learning to meshless methods},}\ }\href@noop {}
  {\bibfield  {journal} {\bibinfo  {journal} {Acta numerica}\ }\textbf
  {\bibinfo {volume} {15}},\ \bibinfo {pages} {543--639} (\bibinfo {year}
  {2006})}\BibitemShut {NoStop}%
\bibitem [{\citenamefont {Chen}\ \emph {et~al.}(2021)\citenamefont {Chen},
  \citenamefont {Hosseini}, \citenamefont {Owhadi},\ and\ \citenamefont
  {Stuart}}]{chen2021solving}%
  \BibitemOpen
  \bibfield  {author} {\bibinfo {author} {\bibfnamefont {Y.}~\bibnamefont
  {Chen}}, \bibinfo {author} {\bibfnamefont {B.}~\bibnamefont {Hosseini}},
  \bibinfo {author} {\bibfnamefont {H.}~\bibnamefont {Owhadi}}, \ and\ \bibinfo
  {author} {\bibfnamefont {A.~M.}\ \bibnamefont {Stuart}},\ }\bibfield  {title}
  {\enquote {\bibinfo {title} {Solving and learning nonlinear {PDEs} with
  {G}aussian processes},}\ }\href@noop {} {\bibfield  {journal} {\bibinfo
  {journal} {Journal of {C}omputational {P}hysics}\ }\textbf {\bibinfo {volume}
  {447}},\ \bibinfo {pages} {110668} (\bibinfo {year} {2021})}\BibitemShut
  {NoStop}%
\bibitem [{\citenamefont {Rowley}\ \emph {et~al.}(2009)\citenamefont {Rowley},
  \citenamefont {Mezi{\'c}}, \citenamefont {Bagheri}, \citenamefont
  {Schlatter},\ and\ \citenamefont {Henningson}}]{RMBSH09}%
  \BibitemOpen
  \bibfield  {author} {\bibinfo {author} {\bibfnamefont {C.~W.}\ \bibnamefont
  {Rowley}}, \bibinfo {author} {\bibfnamefont {I.}~\bibnamefont {Mezi{\'c}}},
  \bibinfo {author} {\bibfnamefont {S.}~\bibnamefont {Bagheri}}, \bibinfo
  {author} {\bibfnamefont {P.}~\bibnamefont {Schlatter}}, \ and\ \bibinfo
  {author} {\bibfnamefont {D.~S.}\ \bibnamefont {Henningson}},\ }\bibfield
  {title} {\enquote {\bibinfo {title} {Spectral analysis of nonlinear flows},}\
  }\href {\doibase 10.1017/S0022112009992059} {\bibfield  {journal} {\bibinfo
  {journal} {Journal of Fluid Mechanics}\ }\textbf {\bibinfo {volume} {641}},\
  \bibinfo {pages} {115--127} (\bibinfo {year} {2009})}\BibitemShut {NoStop}%
\bibitem [{\citenamefont {Budi{\v s}i{\'c}}, \citenamefont {Mohr},\ and\
  \citenamefont {Mezi{\'c}}(2012)}]{BMM12}%
  \BibitemOpen
  \bibfield  {author} {\bibinfo {author} {\bibfnamefont {M.}~\bibnamefont
  {Budi{\v s}i{\'c}}}, \bibinfo {author} {\bibfnamefont {R.}~\bibnamefont
  {Mohr}}, \ and\ \bibinfo {author} {\bibfnamefont {I.}~\bibnamefont
  {Mezi{\'c}}},\ }\bibfield  {title} {\enquote {\bibinfo {title} {Applied
  {K}oopmanism},}\ }\href {\doibase 10.1063/1.4772195} {\bibfield  {journal}
  {\bibinfo  {journal} {Chaos: An Interdisciplinary Journal of Nonlinear
  Science}\ }\textbf {\bibinfo {volume} {22}} (\bibinfo {year} {2012}),\
  10.1063/1.4772195}\BibitemShut {NoStop}%
\bibitem [{\citenamefont {Sch{\"o}lkopf}\ and\ \citenamefont
  {Smola}(2001)}]{Schoe01}%
  \BibitemOpen
  \bibfield  {author} {\bibinfo {author} {\bibfnamefont {B.}~\bibnamefont
  {Sch{\"o}lkopf}}\ and\ \bibinfo {author} {\bibfnamefont {A.~J.}\ \bibnamefont
  {Smola}},\ }\href@noop {} {\emph {\bibinfo {title} {Learning with Kernels:
  Support Vector Machines, Regularization, Optimization and Beyond}}}\
  (\bibinfo  {publisher} {MIT press},\ \bibinfo {address} {Cambridge, USA},\
  \bibinfo {year} {2001})\BibitemShut {NoStop}%
\bibitem [{\citenamefont {Steinwart}\ and\ \citenamefont
  {Christmann}(2008)}]{Steinwart2008:SVM}%
  \BibitemOpen
  \bibfield  {author} {\bibinfo {author} {\bibfnamefont {I.}~\bibnamefont
  {Steinwart}}\ and\ \bibinfo {author} {\bibfnamefont {A.}~\bibnamefont
  {Christmann}},\ }\href@noop {} {\emph {\bibinfo {title} {Support Vector
  Machines}}},\ \bibinfo {edition} {1st}\ ed.\ (\bibinfo  {publisher}
  {Springer},\ \bibinfo {address} {New York},\ \bibinfo {year}
  {2008})\BibitemShut {NoStop}%
\bibitem [{\citenamefont {Aronszajn}(1950)}]{Aronszajn50}%
  \BibitemOpen
  \bibfield  {author} {\bibinfo {author} {\bibfnamefont {N.}~\bibnamefont
  {Aronszajn}},\ }\bibfield  {title} {\enquote {\bibinfo {title} {Theory of
  reproducing kernels},}\ }\href@noop {} {\bibfield  {journal} {\bibinfo
  {journal} {Transactions of the American Mathematical Society}\ }\textbf
  {\bibinfo {volume} {68}},\ \bibinfo {pages} {337--404} (\bibinfo {year}
  {1950})}\BibitemShut {NoStop}%
\bibitem [{\citenamefont {Kimeldorf}\ and\ \citenamefont
  {Wahba}(1971)}]{kimeldorf1971some}%
  \BibitemOpen
  \bibfield  {author} {\bibinfo {author} {\bibfnamefont {G.}~\bibnamefont
  {Kimeldorf}}\ and\ \bibinfo {author} {\bibfnamefont {G.}~\bibnamefont
  {Wahba}},\ }\bibfield  {title} {\enquote {\bibinfo {title} {Some results on
  {T}chebycheffian spline functions},}\ }\href@noop {} {\bibfield  {journal}
  {\bibinfo  {journal} {Journal of Mathematical Analysis and Applications}\
  }\textbf {\bibinfo {volume} {33}},\ \bibinfo {pages} {82--95} (\bibinfo
  {year} {1971})}\BibitemShut {NoStop}%
\bibitem [{\citenamefont {Sch{\"o}lkopf}, \citenamefont {Herbrich},\ and\
  \citenamefont {Smola}(2001)}]{scholkopf2001generalized}%
  \BibitemOpen
  \bibfield  {author} {\bibinfo {author} {\bibfnamefont {B.}~\bibnamefont
  {Sch{\"o}lkopf}}, \bibinfo {author} {\bibfnamefont {R.}~\bibnamefont
  {Herbrich}}, \ and\ \bibinfo {author} {\bibfnamefont {A.~J.}\ \bibnamefont
  {Smola}},\ }\bibfield  {title} {\enquote {\bibinfo {title} {A generalized
  representer theorem},}\ }in\ \href@noop {} {\emph {\bibinfo {booktitle}
  {International conference on computational learning theory}}}\ (\bibinfo
  {organization} {Springer},\ \bibinfo {year} {2001})\ pp.\ \bibinfo {pages}
  {416--426}\BibitemShut {NoStop}%
\bibitem [{\citenamefont {Hoerl}\ and\ \citenamefont
  {Kennard}(1970)}]{hoerl1970ridge}%
  \BibitemOpen
  \bibfield  {author} {\bibinfo {author} {\bibfnamefont {A.~E.}\ \bibnamefont
  {Hoerl}}\ and\ \bibinfo {author} {\bibfnamefont {R.~W.}\ \bibnamefont
  {Kennard}},\ }\bibfield  {title} {\enquote {\bibinfo {title} {Ridge
  regression: {B}iased estimation for nonorthogonal problems},}\ }\href@noop {}
  {\bibfield  {journal} {\bibinfo  {journal} {Technometrics}\ }\textbf
  {\bibinfo {volume} {12}},\ \bibinfo {pages} {55--67} (\bibinfo {year}
  {1970})}\BibitemShut {NoStop}%
\bibitem [{\citenamefont {Golub}\ and\ \citenamefont
  {Pereyra}(1973)}]{golub1973differentiation}%
  \BibitemOpen
  \bibfield  {author} {\bibinfo {author} {\bibfnamefont {G.~H.}\ \bibnamefont
  {Golub}}\ and\ \bibinfo {author} {\bibfnamefont {V.}~\bibnamefont
  {Pereyra}},\ }\bibfield  {title} {\enquote {\bibinfo {title} {The
  differentiation of pseudo-inverses and nonlinear least squares problems whose
  variables separate},}\ }\href@noop {} {\bibfield  {journal} {\bibinfo
  {journal} {SIAM Journal on numerical analysis}\ }\textbf {\bibinfo {volume}
  {10}},\ \bibinfo {pages} {413--432} (\bibinfo {year} {1973})}\BibitemShut
  {NoStop}%
\bibitem [{\citenamefont {Golub}\ and\ \citenamefont
  {Pereyra}(2003)}]{golub2003separable}%
  \BibitemOpen
  \bibfield  {author} {\bibinfo {author} {\bibfnamefont {G.}~\bibnamefont
  {Golub}}\ and\ \bibinfo {author} {\bibfnamefont {V.}~\bibnamefont
  {Pereyra}},\ }\bibfield  {title} {\enquote {\bibinfo {title} {Separable
  nonlinear least squares: the variable projection method and its
  applications},}\ }\href@noop {} {\bibfield  {journal} {\bibinfo  {journal}
  {Inverse problems}\ }\textbf {\bibinfo {volume} {19}},\ \bibinfo {pages}
  {R1--R26} (\bibinfo {year} {2003})}\BibitemShut {NoStop}%
\bibitem [{\citenamefont {Liu}\ and\ \citenamefont
  {Nocedal}(1989)}]{liu1989limited}%
  \BibitemOpen
  \bibfield  {author} {\bibinfo {author} {\bibfnamefont {D.~C.}\ \bibnamefont
  {Liu}}\ and\ \bibinfo {author} {\bibfnamefont {J.}~\bibnamefont {Nocedal}},\
  }\bibfield  {title} {\enquote {\bibinfo {title} {On the limited memory {BFGS}
  method for large scale optimization},}\ }\href@noop {} {\bibfield  {journal}
  {\bibinfo  {journal} {Mathematical programming}\ }\textbf {\bibinfo {volume}
  {45}},\ \bibinfo {pages} {503--528} (\bibinfo {year} {1989})}\BibitemShut
  {NoStop}%
\bibitem [{\citenamefont {DeGennaro}\ and\ \citenamefont
  {Urban}(2019)}]{degennaro2019scalable}%
  \BibitemOpen
  \bibfield  {author} {\bibinfo {author} {\bibfnamefont {A.~M.}\ \bibnamefont
  {DeGennaro}}\ and\ \bibinfo {author} {\bibfnamefont {N.~M.}\ \bibnamefont
  {Urban}},\ }\bibfield  {title} {\enquote {\bibinfo {title} {Scalable extended
  dynamic mode decomposition using random kernel approximation},}\ }\href@noop
  {} {\bibfield  {journal} {\bibinfo  {journal} {SIAM Journal on Scientific
  Computing}\ }\textbf {\bibinfo {volume} {41}},\ \bibinfo {pages}
  {A1482--A1499} (\bibinfo {year} {2019})}\BibitemShut {NoStop}%
\bibitem [{\citenamefont {N{\"u}ske}\ and\ \citenamefont {Klus}(2023)}]{NK23}%
  \BibitemOpen
  \bibfield  {author} {\bibinfo {author} {\bibfnamefont {F.}~\bibnamefont
  {N{\"u}ske}}\ and\ \bibinfo {author} {\bibfnamefont {S.}~\bibnamefont
  {Klus}},\ }\bibfield  {title} {\enquote {\bibinfo {title} {Efficient
  approximation of molecular kinetics using random {F}ourier features},}\
  }\href@noop {} {\bibfield  {journal} {\bibinfo  {journal} {The Journal of
  Chemical Physics}\ }\textbf {\bibinfo {volume} {159}} (\bibinfo {year}
  {2023})}\BibitemShut {NoStop}%
\bibitem [{\citenamefont {Sobol}(1967)}]{sobol1967distribution}%
  \BibitemOpen
  \bibfield  {author} {\bibinfo {author} {\bibfnamefont {I.~M.}\ \bibnamefont
  {Sobol}},\ }\bibfield  {title} {\enquote {\bibinfo {title} {Distribution of
  points in a cube and approximate evaluation of integrals},}\ }\href@noop {}
  {\bibfield  {journal} {\bibinfo  {journal} {USSR Computational Mathematics
  and Mathematical Physics}\ }\textbf {\bibinfo {volume} {7}},\ \bibinfo
  {pages} {86--112} (\bibinfo {year} {1967})}\BibitemShut {NoStop}%
\bibitem [{\citenamefont {Reynolds}\ \emph {et~al.}(2002)\citenamefont
  {Reynolds}, \citenamefont {Rayner}, \citenamefont {Smith}, \citenamefont
  {Stokes},\ and\ \citenamefont {Wang}}]{reynolds2002improved}%
  \BibitemOpen
  \bibfield  {author} {\bibinfo {author} {\bibfnamefont {R.~W.}\ \bibnamefont
  {Reynolds}}, \bibinfo {author} {\bibfnamefont {N.~A.}\ \bibnamefont
  {Rayner}}, \bibinfo {author} {\bibfnamefont {T.~M.}\ \bibnamefont {Smith}},
  \bibinfo {author} {\bibfnamefont {D.~C.}\ \bibnamefont {Stokes}}, \ and\
  \bibinfo {author} {\bibfnamefont {W.}~\bibnamefont {Wang}},\ }\bibfield
  {title} {\enquote {\bibinfo {title} {An improved in situ and satellite {SST}
  analysis for climate},}\ }\href@noop {} {\bibfield  {journal} {\bibinfo
  {journal} {Journal of climate}\ }\textbf {\bibinfo {volume} {15}},\ \bibinfo
  {pages} {1609--1625} (\bibinfo {year} {2002})}\BibitemShut {NoStop}%
\bibitem [{\citenamefont {Gao}, \citenamefont {Williams},\ and\ \citenamefont
  {Kutz}(2025)}]{gao2025sparse_dataset}%
  \BibitemOpen
  \bibfield  {author} {\bibinfo {author} {\bibfnamefont {L.~M.}\ \bibnamefont
  {Gao}}, \bibinfo {author} {\bibfnamefont {J.}~\bibnamefont {Williams}}, \
  and\ \bibinfo {author} {\bibfnamefont {N.}~\bibnamefont {Kutz}},\ }\href@noop
  {} {\enquote {\bibinfo {title} {Sparse identification of nonlinear dynamics
  and {K}oopman operators with shallow recurrent decoder networks
  [datasets]},}\ } (\bibinfo {year} {2025})\BibitemShut {NoStop}%
\bibitem [{\citenamefont {Gao}, \citenamefont {Williams},\ and\ \citenamefont
  {Kutz}(2026)}]{gao2026sparse}%
  \BibitemOpen
  \bibfield  {author} {\bibinfo {author} {\bibfnamefont {M.~L.}\ \bibnamefont
  {Gao}}, \bibinfo {author} {\bibfnamefont {J.~P.}\ \bibnamefont {Williams}}, \
  and\ \bibinfo {author} {\bibfnamefont {J.~N.}\ \bibnamefont {Kutz}},\
  }\bibfield  {title} {\enquote {\bibinfo {title} {Sparse identification of
  nonlinear dynamics and {K}oopman operators with shallow recurrent decoder
  networks},}\ }\href@noop {} {\bibfield  {journal} {\bibinfo  {journal}
  {Proceedings of the National Academy of Sciences}\ }\textbf {\bibinfo
  {volume} {123}},\ \bibinfo {pages} {e2508144123} (\bibinfo {year}
  {2026})}\BibitemShut {NoStop}%
\bibitem [{\citenamefont {Brunton}, \citenamefont {Proctor},\ and\
  \citenamefont {Kutz}(2016)}]{BPK16}%
  \BibitemOpen
  \bibfield  {author} {\bibinfo {author} {\bibfnamefont {S.~L.}\ \bibnamefont
  {Brunton}}, \bibinfo {author} {\bibfnamefont {J.~L.}\ \bibnamefont
  {Proctor}}, \ and\ \bibinfo {author} {\bibfnamefont {J.~N.}\ \bibnamefont
  {Kutz}},\ }\bibfield  {title} {\enquote {\bibinfo {title} {Discovering
  governing equations from data by sparse identification of nonlinear dynamical
  systems},}\ }\href@noop {} {\bibfield  {journal} {\bibinfo  {journal}
  {Proceedings of the National Academy of Sciences}\ }\textbf {\bibinfo
  {volume} {113}},\ \bibinfo {pages} {3932--3937} (\bibinfo {year}
  {2016})}\BibitemShut {NoStop}%
\bibitem [{\citenamefont {Williams}, \citenamefont {Zahn},\ and\ \citenamefont
  {Kutz}(2024)}]{williams2024sensing}%
  \BibitemOpen
  \bibfield  {author} {\bibinfo {author} {\bibfnamefont {J.~P.}\ \bibnamefont
  {Williams}}, \bibinfo {author} {\bibfnamefont {O.}~\bibnamefont {Zahn}}, \
  and\ \bibinfo {author} {\bibfnamefont {J.~N.}\ \bibnamefont {Kutz}},\
  }\bibfield  {title} {\enquote {\bibinfo {title} {Sensing with shallow
  recurrent decoder networks},}\ }\href {\doibase 10.1098/rspa.2024.0054}
  {\bibfield  {journal} {\bibinfo  {journal} {Proceedings of the Royal Society
  A: Mathematical, Physical and Engineering Sciences}\ }\textbf {\bibinfo
  {volume} {480}},\ \bibinfo {pages} {20240054} (\bibinfo {year}
  {2024})}\BibitemShut {NoStop}%
\bibitem [{\citenamefont {Batlle}\ \emph {et~al.}(2024)\citenamefont {Batlle},
  \citenamefont {Darcy}, \citenamefont {Hosseini},\ and\ \citenamefont
  {Owhadi}}]{batlle2024kernel}%
  \BibitemOpen
  \bibfield  {author} {\bibinfo {author} {\bibfnamefont {P.}~\bibnamefont
  {Batlle}}, \bibinfo {author} {\bibfnamefont {M.}~\bibnamefont {Darcy}},
  \bibinfo {author} {\bibfnamefont {B.}~\bibnamefont {Hosseini}}, \ and\
  \bibinfo {author} {\bibfnamefont {H.}~\bibnamefont {Owhadi}},\ }\bibfield
  {title} {\enquote {\bibinfo {title} {Kernel methods are competitive for
  operator learning},}\ }\href@noop {} {\bibfield  {journal} {\bibinfo
  {journal} {Journal of Computational Physics}\ }\textbf {\bibinfo {volume}
  {496}},\ \bibinfo {pages} {112549} (\bibinfo {year} {2024})}\BibitemShut
  {NoStop}%
\bibitem [{\citenamefont {Wu}\ and\ \citenamefont {No{\'e}}(2020)}]{WuNo20}%
  \BibitemOpen
  \bibfield  {author} {\bibinfo {author} {\bibfnamefont {H.}~\bibnamefont
  {Wu}}\ and\ \bibinfo {author} {\bibfnamefont {F.}~\bibnamefont {No{\'e}}},\
  }\bibfield  {title} {\enquote {\bibinfo {title} {Variational approach for
  learning {M}arkov processes from time series data},}\ }\href {\doibase
  10.1007/s00332-019-09567-y} {\bibfield  {journal} {\bibinfo  {journal}
  {Journal of Nonlinear Science}\ }\textbf {\bibinfo {volume} {30}},\ \bibinfo
  {pages} {33--66} (\bibinfo {year} {2020})}\BibitemShut {NoStop}%
\bibitem [{\citenamefont {Sch\"utte}, \citenamefont {Klus},\ and\ \citenamefont
  {Hartmann}(2023)}]{SKH23}%
  \BibitemOpen
  \bibfield  {author} {\bibinfo {author} {\bibfnamefont {C.}~\bibnamefont
  {Sch\"utte}}, \bibinfo {author} {\bibfnamefont {S.}~\bibnamefont {Klus}}, \
  and\ \bibinfo {author} {\bibfnamefont {C.}~\bibnamefont {Hartmann}},\
  }\bibfield  {title} {\enquote {\bibinfo {title} {Overcoming the timescale
  barrier in molecular dynamics: {T}ransfer operators, variational principles
  and machine learning},}\ }\href {\doibase 10.1017/S0962492923000016}
  {\bibfield  {journal} {\bibinfo  {journal} {Acta Numerica}\ }\textbf
  {\bibinfo {volume} {32}},\ \bibinfo {pages} {517–673} (\bibinfo {year}
  {2023})}\BibitemShut {NoStop}%
\bibitem [{\citenamefont {Korda}\ and\ \citenamefont
  {Mezi{\'c}}(2018)}]{korda2018convergence}%
  \BibitemOpen
  \bibfield  {author} {\bibinfo {author} {\bibfnamefont {M.}~\bibnamefont
  {Korda}}\ and\ \bibinfo {author} {\bibfnamefont {I.}~\bibnamefont
  {Mezi{\'c}}},\ }\bibfield  {title} {\enquote {\bibinfo {title} {On
  convergence of extended dynamic mode decomposition to the {K}oopman
  operator},}\ }\href@noop {} {\bibfield  {journal} {\bibinfo  {journal}
  {Journal of Nonlinear Science}\ }\textbf {\bibinfo {volume} {28}},\ \bibinfo
  {pages} {687--710} (\bibinfo {year} {2018})}\BibitemShut {NoStop}%
\bibitem [{\citenamefont {Philipp}\ \emph {et~al.}(2025)\citenamefont
  {Philipp}, \citenamefont {Schaller}, \citenamefont {Worthmann}, \citenamefont
  {Peitz},\ and\ \citenamefont {N{\"u}ske}}]{philipp2025error}%
  \BibitemOpen
  \bibfield  {author} {\bibinfo {author} {\bibfnamefont {F.~M.}\ \bibnamefont
  {Philipp}}, \bibinfo {author} {\bibfnamefont {M.}~\bibnamefont {Schaller}},
  \bibinfo {author} {\bibfnamefont {K.}~\bibnamefont {Worthmann}}, \bibinfo
  {author} {\bibfnamefont {S.}~\bibnamefont {Peitz}}, \ and\ \bibinfo {author}
  {\bibfnamefont {F.}~\bibnamefont {N{\"u}ske}},\ }\bibfield  {title} {\enquote
  {\bibinfo {title} {Error analysis of kernel {EDMD} for prediction and control
  in the {K}oopman framework},}\ }\href@noop {} {\bibfield  {journal} {\bibinfo
   {journal} {Journal of Nonlinear Science}\ }\textbf {\bibinfo {volume}
  {35}},\ \bibinfo {pages} {92} (\bibinfo {year} {2025})}\BibitemShut {NoStop}%
\bibitem [{\citenamefont {K{\"o}hne}\ \emph {et~al.}(2025)\citenamefont
  {K{\"o}hne}, \citenamefont {Philipp}, \citenamefont {Schaller}, \citenamefont
  {Schiela},\ and\ \citenamefont {Worthmann}}]{kohne2025error}%
  \BibitemOpen
  \bibfield  {author} {\bibinfo {author} {\bibfnamefont {F.}~\bibnamefont
  {K{\"o}hne}}, \bibinfo {author} {\bibfnamefont {F.~M.}\ \bibnamefont
  {Philipp}}, \bibinfo {author} {\bibfnamefont {M.}~\bibnamefont {Schaller}},
  \bibinfo {author} {\bibfnamefont {A.}~\bibnamefont {Schiela}}, \ and\
  \bibinfo {author} {\bibfnamefont {K.}~\bibnamefont {Worthmann}},\ }\bibfield
  {title} {\enquote {\bibinfo {title} {{$L^\infty$}-error bounds for
  approximations of the {K}oopman operator by kernel extended dynamic mode
  decomposition},}\ }\href@noop {} {\bibfield  {journal} {\bibinfo  {journal}
  {SIAM journal on applied dynamical systems}\ }\textbf {\bibinfo {volume}
  {24}},\ \bibinfo {pages} {501--529} (\bibinfo {year} {2025})}\BibitemShut
  {NoStop}%
\bibitem [{\citenamefont {Batlle}\ \emph {et~al.}(2025)\citenamefont {Batlle},
  \citenamefont {Chen}, \citenamefont {Hosseini}, \citenamefont {Owhadi},\ and\
  \citenamefont {Stuart}}]{batlle2025error}%
  \BibitemOpen
  \bibfield  {author} {\bibinfo {author} {\bibfnamefont {P.}~\bibnamefont
  {Batlle}}, \bibinfo {author} {\bibfnamefont {Y.}~\bibnamefont {Chen}},
  \bibinfo {author} {\bibfnamefont {B.}~\bibnamefont {Hosseini}}, \bibinfo
  {author} {\bibfnamefont {H.}~\bibnamefont {Owhadi}}, \ and\ \bibinfo {author}
  {\bibfnamefont {A.~M.}\ \bibnamefont {Stuart}},\ }\bibfield  {title}
  {\enquote {\bibinfo {title} {Error analysis of kernel/{GP} methods for
  nonlinear and parametric {PDEs}},}\ }\href@noop {} {\bibfield  {journal}
  {\bibinfo  {journal} {Journal of Computational Physics}\ }\textbf {\bibinfo
  {volume} {520}},\ \bibinfo {pages} {113488} (\bibinfo {year}
  {2025})}\BibitemShut {NoStop}%
\bibitem [{\citenamefont {Sch{\"o}lkopf}, \citenamefont {Smola},\ and\
  \citenamefont {M{\"u}ller}(1998)}]{scholkopf1998nonlinear}%
  \BibitemOpen
  \bibfield  {author} {\bibinfo {author} {\bibfnamefont {B.}~\bibnamefont
  {Sch{\"o}lkopf}}, \bibinfo {author} {\bibfnamefont {A.}~\bibnamefont
  {Smola}}, \ and\ \bibinfo {author} {\bibfnamefont {K.-R.}\ \bibnamefont
  {M{\"u}ller}},\ }\bibfield  {title} {\enquote {\bibinfo {title} {Nonlinear
  component analysis as a kernel eigenvalue problem},}\ }\href@noop {}
  {\bibfield  {journal} {\bibinfo  {journal} {Neural computation}\ }\textbf
  {\bibinfo {volume} {10}},\ \bibinfo {pages} {1299--1319} (\bibinfo {year}
  {1998})}\BibitemShut {NoStop}%
\end{thebibliography}%
\end{document}